\pdfoutput=1
\documentclass[11pt]{article}

\usepackage[]{acl}

\usepackage{times}
\usepackage{latexsym}

\usepackage[T1]{fontenc}
\usepackage[utf8]{inputenc}

\usepackage{microtype}

\usepackage{booktabs, tabularx}
\usepackage{amssymb}
\usepackage{subfig}

\usepackage{pgfplots}
\usepgfplotslibrary{fillbetween}
\usetikzlibrary{patterns}
\usetikzlibrary{pgfplots.groupplots}
\pgfplotsset{compat=1.17}
\usepackage{lscape}
\usepackage{multirow}
\usepackage{dblfloatfix}
\usepackage{footnote}
\usepackage{graphicx}
\usepackage{tikz}

\newcommand\blfootnote[1]{%
  \begingroup
  \renewcommand\thefootnote{}\footnote{#1}%
  \addtocounter{footnote}{-1}%
  \endgroup
}
\newcommand{\mathboldface}[1]{\boldsymbol{#1}}
\newcommand{\bm}[1]{\mathboldface{#1}}
\newcommand{\bgfg}[1]{\tikz[baseline=(X.base)]{\node(X)[rectangle, fill=gray!15, rounded corners, text height=1.4ex,text depth=-0.5ex,draw=white]{#1};}}

\usepackage{amsmath, mathtools}
\usepackage{amssymb}

\providecommand{\methodName}{GlobEnc}

\usepackage{hyphenat}
\DeclareCaptionTextFormat{new}{\nohyphens{#1}}

\title{\methodName: Quantifying Global Token Attribution by Incorporating the Whole Encoder Layer in Transformers}

\author{
    Ali Modarressi$^{1\star}$ ~ Mohsen Fayyaz$^{2\star}$ ~ Yadollah Yaghoobzadeh$^{2}$  ~ \textbf{Mohammad Taher Pilehvar$^{3}$} \\
    $^1$ Iran University of Science and Technology, Iran ~
    $^2$ University of Tehran, Iran \\
    $^3$ Tehran Institute for Advanced Studies, Khatam University, Iran \\
    \texttt{m\_modarressi@comp.iust.ac.ir}\\
    \texttt{\{mohsen.fayyaz77, y.yaghoobzadeh\}@ut.ac.ir}\\
    \texttt{mp792@cam.ac.uk}
  }

\begin{document}
\maketitle
\begin{abstract}
There has been a growing interest in interpreting the underlying dynamics of Transformers.
While self-attention patterns were initially deemed as the primary option, recent studies have shown that integrating other components can yield more accurate explanations.
This paper introduces a novel token attribution analysis method that incorporates all the components in the encoder block and aggregates this throughout layers. 
Through extensive quantitative and qualitative experiments, we demonstrate that our method can produce faithful and meaningful global token attributions.
Our experiments reveal that incorporating almost every encoder component results in increasingly more accurate analysis in both local (single layer) and global (the whole model) settings. 
Our global attribution analysis significantly outperforms previous methods on various tasks regarding correlation with gradient-based saliency scores.
Our code is freely available at \href{https://github.com/mohsenfayyaz/GlobEnc}{https://github.com/mohsenfayyaz/GlobEnc}.

\blfootnote{$^\star$ Equal contribution.}
\end{abstract}

\section{Introduction}
The stellar performance of Transformers \cite{attention-is-all-you-need} has garnered a lot of attention to analyzing the reasons behind their effectiveness. The self-attention mechanism has been one of the main areas of focus \cite{clark-etal-2019-bert-look, kovaleva-etal-2019-revealing, reif2019visualizing, bert-track-syntactic-dep}. 
However, there have been debates on whether raw attention weights are reliable anchors for explaining model's behavior or not \cite{wiegreffe-pinter-2019-attention, serrano-smith-2019-attention, jain-wallace-2019-attention}.
Recently, it was shown that incorporating vector norms should be an indispensable part of any attention-based analysis\footnote{We also have shown the unreliability of weights due to norm disparities in probing studies \cite{fayyaz-etal-2021-models}.} \cite{kobayashi-etal-2020-attention-norm, kobayashi-etal-2021-incorporating-residual}.
However, these norm-based studies incorporate only the attention block into their analysis, whereas Transformer encoder layer is composed of more components.

\newcommand\qualclip{40}
\begin{figure}[t]
\centering
    \subfloat{
        % left bottom right top
        \includegraphics[width=0.22\textwidth, trim=8 9 36 50, clip] {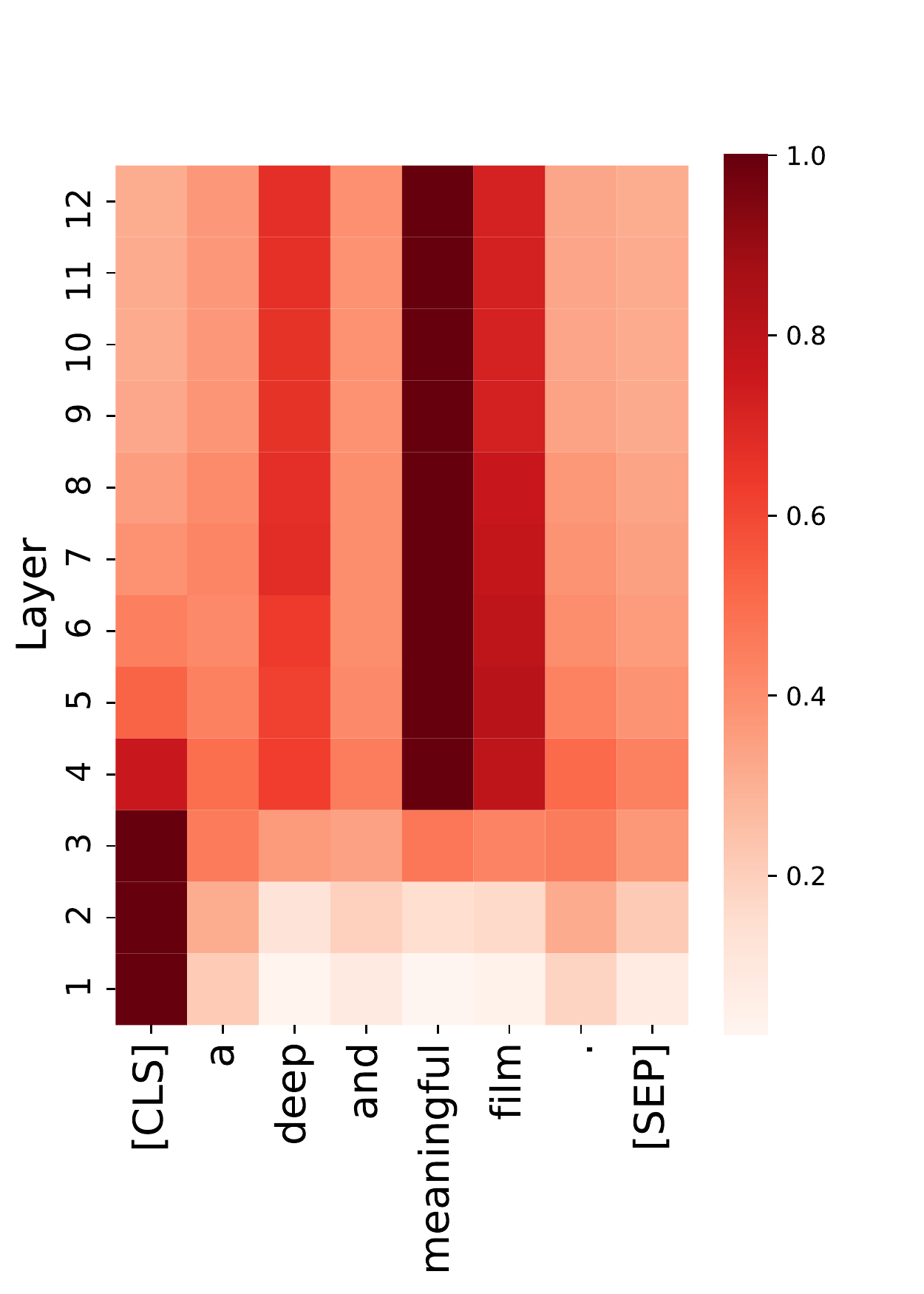}
    }
    \subfloat{
        \includegraphics[width=0.22\textwidth, trim=8 9 36 50, clip] {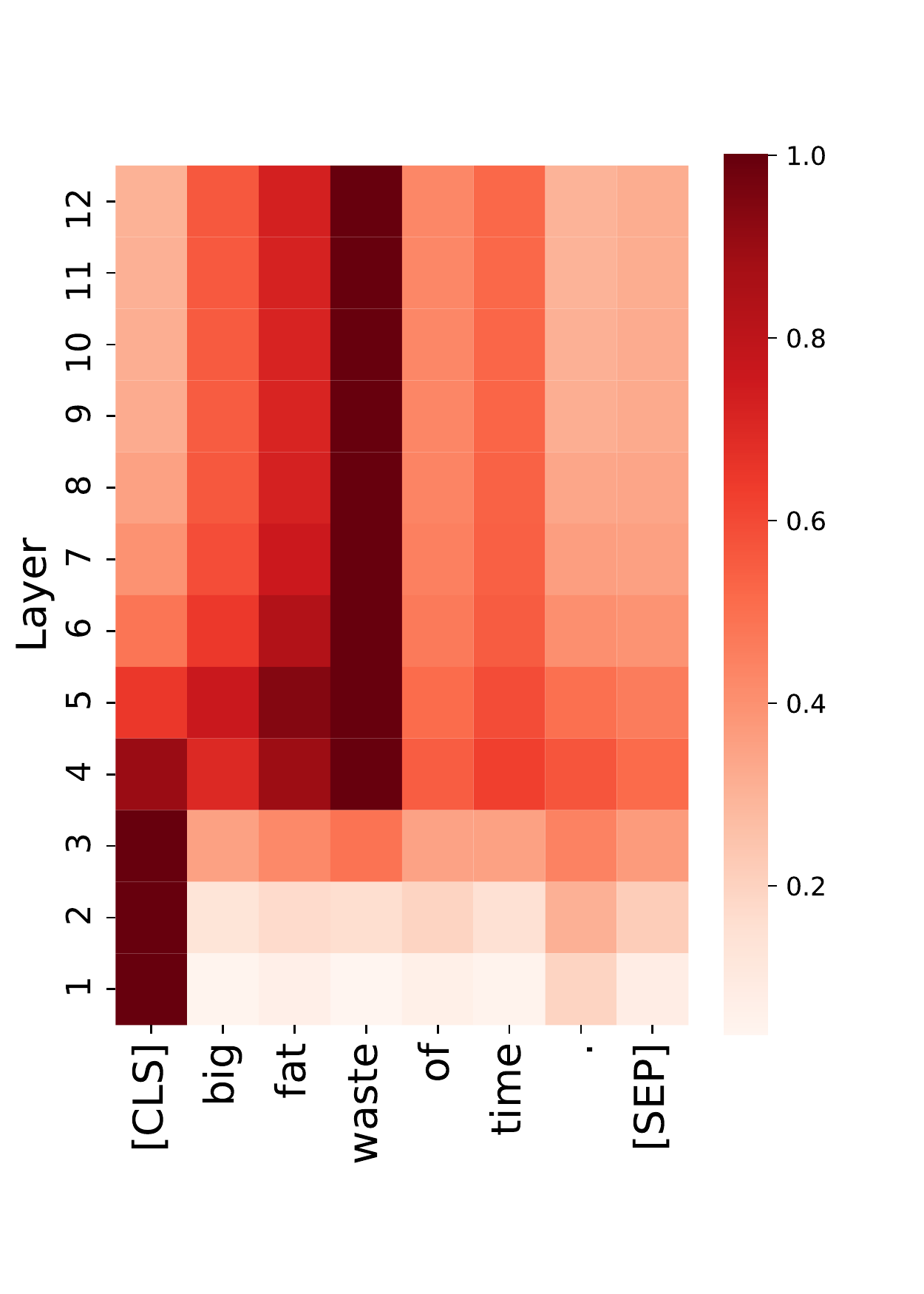}
    }
    
    \caption{
    Aggregated attribution maps ($\mathcal{N_\textsc{Enc}}$) for the \textsc{[CLS]} token for fine-tuned BERT on SST2 dataset (sentiment analysis). Our method (\methodName) is able to accurately quantify the global token attribution of the model.
    }
    \label{fig:sst2_qualitative}
\end{figure}

Another limitation of the existing analysis techniques is that they are usually constrained to the analysis of single layer attributions.
In order to expand the analysis to multi-layered encoder-based models in their entirety,
an aggregation technique has to be employed. 
\citet{abnar-zuidema-2020-quantifying} proposed two aggregation methods, \emph{rollout} and \emph{max-flow}, which combine raw attention weights across layers.
Despite showing the outcome of their method to be faithful to a model's inner workings in specific cases, the final results are still unsatisfactory on a wide range of fine-tuned models.

Additionally, gradient-based alternatives \cite{Simonyan2014DeepIC, kindermans2016investigating, li-etal-2016-visualizing} have been argued to provide a more robust basis for token attribution analysis \cite{atanasova-etal-2020-diagnostic, Brunner2020On, pascual-etal-2021-telling-full-story}.
Nonetheless, the gradient-based alternatives have not been able to fully replace attention-based counterparts, mainly due to their high computational intensity.

In this paper, we propose a new global token attribution analysis method (\methodName) which is based on the encoder layer's output. In \methodName, the second layer normalization is also included in the norm-based analysis of each encoder layer. To aggregate attributions over all layers, we applied a modified attention rollout technique, returning global scores.

Through extensive experiments and comparing the global attribution with the input token attributions obtained by gradient-based saliency scores, we show that our method produces faithful and meaningful results (Figure~\ref{fig:sst2_qualitative}).
Our evaluations on models with distinct pre-training objectives and sizes \cite{devlin-etal-2019-bert, clark2020electra} show high correlations with gradient-based methods in global settings.
Furthermore, with comparative studies on each aspect of \methodName~, we find that: 
(i) norm-based methods achieve higher correlations than weight-based methods;
(ii) incorporating residual connections plays an essential role in token attribution;
(iii) considering the two layer normalizations  improve our analysis only if coupled together; and
(iv) aggregation across layers is crucial for an accurate whole-model attribution analysis.

In summary, our main contributions are:
\begin{itemize}
    \itemsep0em 
    \item We expand the scope of analysis from attention block in Transformers to the whole encoder.
    \item Our method significantly improves over existing techniques for quantifying global token attributions.
    \item We qualitatively demonstrate that the attributions obtained by our method are plausibly interpretable.
\end{itemize}

\section{Background}
\label{sec:background}
\begin{figure}[t]
\centering
    % left bottom right top
    \includegraphics[width=0.48\textwidth, trim=0 100 410 0, clip]{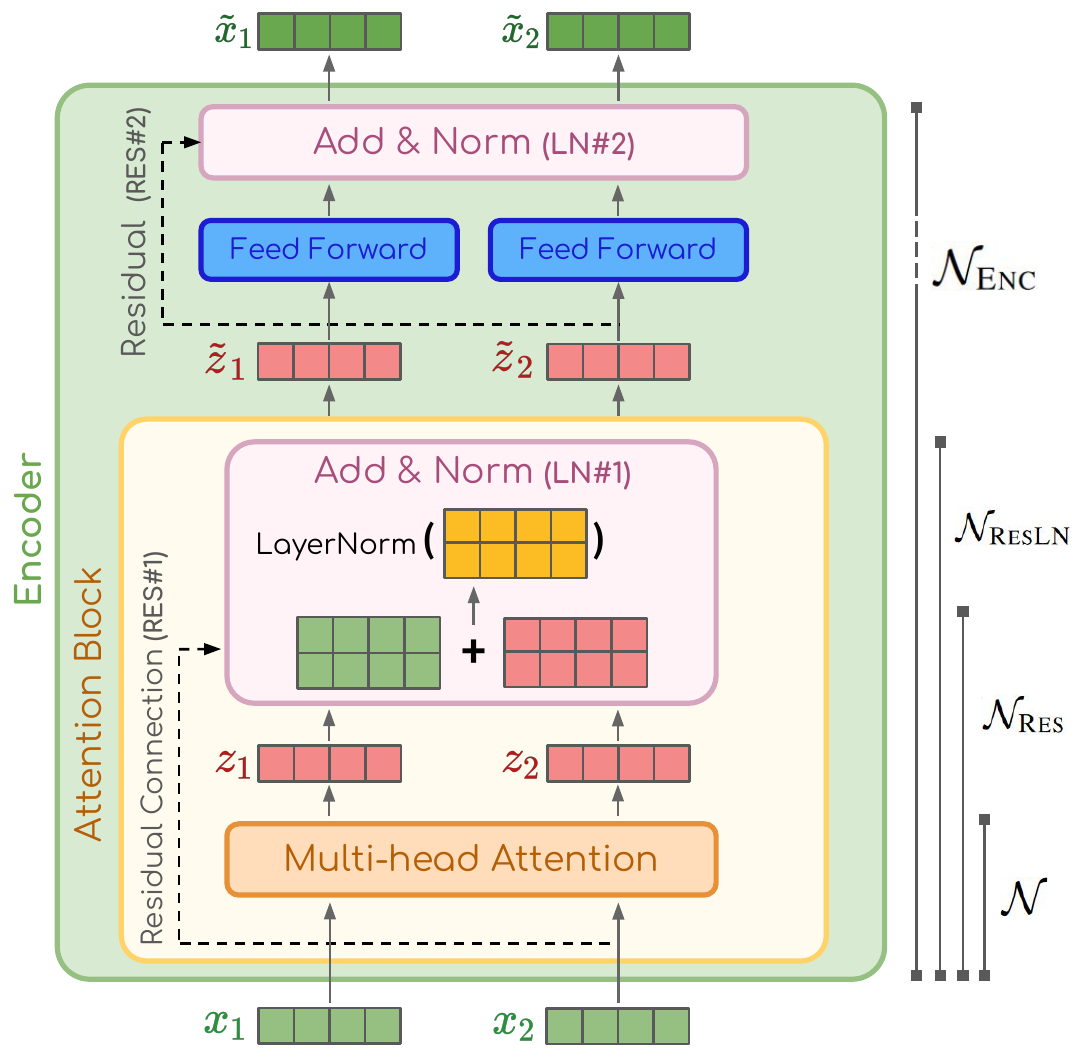}
    \caption{The internal structure of a Transformer encoder layer. We show on the diagram the components that are incorporated by each token attribution analysis method. Our method incorporates the whole encoder ($\mathcal{N_\textsc{Enc}}$) 
    except for the direct effect of the fully connected feed-forward module.
    Diagram inspired by \citet{alammar_j}.
    }
    \label{fig:encoder_diagram}
\end{figure}

In encoder-based language models (such as BERT), a Transformer encoder layer is composed of several components (Figure \ref{fig:encoder_diagram}).
The core component of the encoder is the self-attention mechanism \cite{attention-is-all-you-need}, which is responsible for the information mixture of a sequence of token representations ($\bm{x}_1, ..., \bm{x}_n$). 
Each self-attention head computes a set of attention weights $\bm A^h=\{\alpha_{i,j}^h | 1 \leq i, j \leq n \}$, where $\alpha_{i,j}^h$ is the raw attention weight from the $i^\text{th}$ token to the $j^\text{th}$ token in head $h \in \{1,...,H\}$.
Therefore, the output representation ($\bm{z}_i \in \mathbb{R}^d$) for the $i^\text{th}$ token of a multi-head (with $H$ heads) self-attention module is computed by concatenating the heads' outputs followed by a head-mixing $\bm{W_O}$ projection:
\begin{equation}
\label{eq:self_attention_vaswani}
\bm{z}_i = \textsc{Concat}(\bm{z}_i^1, ..., \bm{z}_i^H)\bm{W_O}
\end{equation}
where each head's output vector is generated by performing a weighted sum over the transformed value vectors $\bm{v}(\bm{x}_j) \in \mathbb{R}^{d_v}$:
\begin{equation}
\label{eq:self_attention_vaswani_head}
\bm{z}_i^h = \sum_{j=1}^{n}\alpha_{i,j}^h\bm{v}^h(\bm{x}_j)
\vspace{-3.0ex}
\end{equation}

\paragraph{Norm-based attention.} 
While one may interpret the attention mechanism using the attention weights $\bm A$, \citet{kobayashi-etal-2020-attention-norm} argued that doing so would ignore the norm of the transformed vectors multiplied by the weights, elucidating that the weights are insufficient for interpretation. Their solution enhanced the interpretability of attention weights by incorporating the value vectors $\bm{v}(\bm{x}_j)$ and the following projection  $\bm W_O$. 
By reformulating Equation \ref{eq:self_attention_vaswani}, we can consider $\bm{z}_i$ as a summation over the attentions heads:
\begin{equation}
\label{eq:self_attention_reformulated}
\bm{z}_i = \sum_{h=1}^{H}\sum_{j=1}^{n}\alpha^h_{i,j}\underbrace{\bm{v}^h(\bm{x}_j)\bm W^h_O}_{f^h(\bm{x}_j)}
\end{equation}
Using this reformulation\footnote{
$\bm W^h_O$ is a head-specific slice of the original $\bm W_O$ projection. For more information about the reformulation process, see Appendix C in \citet{kobayashi-etal-2021-incorporating-residual}}, \citeauthor{kobayashi-etal-2020-attention-norm} proposed a \emph{norm-based} token attribution analysis method, 
\bgfg{
$\mathcal{N}:=(||\bm{z}_{i\leftarrow j}||) \in \mathbb{R}^{n\times n}$
}
, to measure each token's contribution in a self-attention module:
\begin{equation}
\label{eq:z_j_to_i}
\bm{z}_{i\leftarrow j} = \sum_{h=1}^{H}\alpha^h_{i,j}f^h(\bm{x}_j)
\end{equation}
They showed that incorporating the magnitude of the transformation function ($f^h(\bm{x})$) is crucial in assessing the input tokens' contribution to the self-attention output.
\paragraph{Residual connections \& Layer Normalizations.}
\citet{kobayashi-etal-2021-incorporating-residual} added the attention block's Layer Normalization (\textbf{LN}\#1)
and Residual connection (\textbf{RES}\#1)
to its prior norm-based analysis to assess the impact of residual connections and layer normalization inside an attention block. 
\bgfg{
$\mathcal{N}_\textsc{Res}:=(||\bm{z}^\texttt{+}_{i\leftarrow j}||) \in \mathbb{R}^{n\times n}$
}
is the analysis method which incorporates the attention block's residual connection.
The input vector $\bm x$ is added to the attribution of each token to itself to incorporate the influence of RES\#1:
\begin{equation}
\label{eq:kobayashi_res}
\bm{z}^\texttt{+}_{i\leftarrow j} =
        \sum_{h=1}^{H}\alpha^h_{i,j}f^h(\bm{x}_i) + \mathbf{1}[i=j]\bm{x}_i
\end{equation}
They proposed a method for decomposing LN\footnote{$\bm{\gamma} \in \mathbb{R}^d$ and $\bm{\beta} \in \mathbb{R}^d$ are the trainable weights of LN. Similar to \citet{kobayashi-etal-2021-incorporating-residual} we ignore $\bm{\beta}$.} into a summation of normalizations:
\begin{equation}
\label{eq:kobayashi_LN_decomp}
\begin{aligned}
\text{LN}(\bm{z}_i^\texttt{+}) &= \sum_{j=1}^{n}g_{\bm{z}^\texttt{+}_i}(\bm{z}_{i\leftarrow j}^\texttt{+}) + \bm \beta\\
g_{\bm{z}_i^\texttt{+}}(\bm{z}^\texttt{+}_{i\leftarrow j}) &:= \frac{\bm{z}^\texttt{+}_{i\leftarrow j} - m(\bm{z}^\texttt{+}_{i\leftarrow j})}{s(\bm{z}^\texttt{+}_i)} \odot \bm\gamma
\end{aligned}
\end{equation}
where $m(.)$ and $s(.)$ are the element-wise mean and standard deviation of the input vector (cf. \S\ref{sec:formulation}).
The decomposition can be applied to the contribution vectors:
\begin{equation}
\label{eq:kobayashi_res_LN}
\bm{\tilde{z}}_{i\leftarrow j} =
        g_{\bm{z}^\texttt{+}_i}(\sum_{h=1}^{H}\alpha^h_{i,j}f^h(\bm{x}_i) + \mathbf{1}[i=j]\bm{x}_i)
\end{equation}
Accordingly, we can compute 
the magnitude
\bgfg{
$\mathcal{N}_\textsc{ResLN}:=(||\bm{\tilde{z}}_{i\leftarrow j}||) \in \mathbb{R}^{n\times n}$
}, which represents the amount of influence of an encoder layer's input token $j$ on its output token $i$. Based on this formulation, a context-mixing ratio could be defined as:
\begin{equation}
\label{eq:mixing_ratio}
r_i = \frac{||\sum_{j=1,j\neq i}^{n}\bm{\tilde{z}}_{i\leftarrow j}||}{||\sum_{j=1,j\neq i}^{n}\bm{\tilde{z}}_{i\leftarrow j}|| + ||\bm{\tilde{z}}_{i\leftarrow i}||}
\end{equation}
Experiments by \citet{kobayashi-etal-2021-incorporating-residual} revealed considerably low $r$ values which indicate the huge impact of the residual connections. In other words, the model tends to preserve token representations more than mixing them with each other.

%%%%%%%%%Proposal
\section{Methodology}
\label{sec:methodology}
Our method for input token attribution analysis has a holistic view and takes into account almost every component within the encoder layer.
To this end, we first extend the norm-based analysis of \citet{kobayashi-etal-2021-incorporating-residual} by incorporating the encoder's output LN\#2. 
We then apply an aggregation technique to combine the information flow throughout all layers.

\paragraph{Encoder layer output $\neq$ Attention block output.}
While the RES\#1 and the LN\#1 from the attention block are included in the analysis of \citet{kobayashi-etal-2021-incorporating-residual}, the subsequent FFN, RES\#2, and output LN\#2 are ignored (see Fig.~\ref{fig:encoder_diagram}). 
Hence, $\mathcal{N_\textsc{ResLN}}$ might not be indicative of the entire encoder layer's function.
To address this issue, we additionally include the encoder layer components from the attention block outputs ($\tilde{z}_{i}$) to the output representations ($\tilde{x}_{i}$).
The output of each encoder ($\tilde{x}_{i}$) is computed as follows:
\begin{equation}
\label{eq:output_eq}
\begin{aligned}
\bm{\tilde{z}}_{i}^\texttt{+} &= \text{FFN}(\bm{\tilde{z}}_{i}) + \bm{\tilde{z}}_{i} \\
\bm{\tilde{x}}_{i} &= \text{LN}(\bm{\tilde{z}}_{i}^\texttt{+}) 
\end{aligned}
\end{equation}
We apply the LN decomposition rule in Eq. \ref{eq:kobayashi_res_LN} to separate the impacts of residual and FFN output:
\begin{equation}
\label{eq:output_eq_LNdecomp}
\bm{\tilde{x}}_{i} = \sum_{j=1}^{n} \Big( g_{\bm{\tilde{z}}_{i}^{\texttt{+}}}(\text{FFN}(\bm{\tilde{z}}_{i\leftarrow j})) + g_{\bm{\tilde{z}}_{i}^{\texttt{+}}}(\bm{\tilde{z}}_{i\leftarrow j})\Big) + \bm \beta
\end{equation}
Given that the activation function between the two fully connected layers in the FFN component is non-linear \citep{attention-is-all-you-need}, a linear decomposition similar to Eq. \ref{eq:kobayashi_res_LN} cannot be derived. 
As a result, we omit FFN's influence on the contribution of each token and instead consider RES\#2, approximating $\bm{\tilde{x}}_{i\leftarrow j}$ as $g_{\bm{\tilde{z}}_{i}^{\texttt{+}}}(\bm{\tilde{z}}_{i\leftarrow j})$. 
Nevertheless, it should be noted that the FFN \emph{still preserves some influence on this new setting due to the presence of $s(\bm{\tilde{z}}_{i}^{\texttt{+}})$ in $g_{\bm{\tilde{z}}_{i}^{\texttt{+}}}(\bm{\tilde{z}}_{i\leftarrow j})$.}
Similarly to Eq.~\ref{eq:kobayashi_res_LN}, we can introduce a more inclusive layerwise analysis method
\bgfg{
$\mathcal{N}_\textsc{Enc}:=(||\bm{\tilde{x}}_{i\leftarrow j}||) \in \mathbb{R}^{n\times n}$
}
from input token $j$ to output token $i$ using:
\begin{equation}
\label{eq:encoder_attr}
\bm{\tilde{x}}_{i\leftarrow j} \approx
        g_{\bm{\tilde{z}}_{i}^{\texttt{+}}}(\bm{\tilde{z}}_{i\leftarrow j})
        = 
        \frac{\bm{\tilde{z}}_{i\leftarrow j} - m(\bm{\tilde{z}}_{i\leftarrow j})}{s(\bm{\tilde{z}}_{i}^{\texttt{+}})} \odot \bm\gamma
\end{equation}

\paragraph{Aggregating multi-layer attention.}
\label{sec:rollout}
To create an aggregated attribution score, \citet{abnar-zuidema-2020-quantifying} proposed describing the model's attentions via modelling the information flow with a directed graph.
They introduced \emph{attention rollout} method, which linearly combines raw attention weights along all available paths in the pairwise attention graph. The attention rollout of layer $\ell$ w.r.t. the inputs is computed recursively as follows:
\begin{equation}
\label{eq:attention_rollout_main}
\begin{aligned}
\tilde{\bm{A}}_{\ell} &=
\begin{dcases}
        \Hat{\bm{A}}_{\ell}\tilde{\bm{A}}_{\ell-1} & \ell > 1\\
        \Hat{\bm{A}}_{\ell} & \ell = 1
    \end{dcases} \\
\end{aligned}
\end{equation}
\begin{equation}
\label{eq:attention_rollout_residual}
\begin{aligned}
\Hat{\bm{A}}_{\ell} &= 0.5\Bar{\bm{A}}_{\ell}+0.5\bm{I}
\end{aligned}
\end{equation}
$\Bar{\bm{A}}_{\ell}$ is the raw attention map averaged across all heads in layer $\ell$. 
This method assumes equal contribution from the residual connection and multi-head attention (See Fig.~\ref{fig:encoder_diagram}). 
Hence, an identity matrix is summed and renormalized, giving $\Hat{\bm{A}}_{\ell}$.

For aggregating the layerwise analysis methods, we use the rollout technique with minor modifications.
As many of the methods already include residual connections, we only use Eq. \ref{eq:attention_rollout_main} (replacing $\Hat{\bm{A}}_{\ell}$ with the desired method's attribution matrix in layer $\ell$) to calculate the rollout of a given method. 
However, for methods that do not assume the residual connection, we define a corresponding ``Fixed'' variation using Eq. \ref{eq:attention_rollout_residual} that incorporates a fixed residual effect ($r_i \approx 0.5$).
\textbf{We refer to our proposed global method\textemdash aggregating the $\mathcal{N_\textsc{Enc}}$ across all layers by the rollout method\textemdash as \emph{\methodName}.}
In what follows we report our experiments, comparing \methodName~ with several other settings.

%%%%%%%%%Experiments
\section{Experiments}
In this section, we introduce the datasets and the token attribution analysis methods used in our evaluations, followed by the experimental setup and results.
\subsection{Datasets}
All analysis methods are evaluated on three different classification tasks. 
To cover sentiment detection tasks we use SST2 \cite{socher-etal-2013-recursive}, MNLI \cite{williams-etal-2018-broad} for Natural Language Inference and Hatexplain \cite{Mathew_Saha_Yimam_Biemann_Goyal_Mukherjee_2021} in hate speech detection.

\subsection{Analysis Methods}
\label{sec:analysis_methods}
We use two categories of explainability approaches in our work: \textbf{\emph{Weight-based}} and \textbf{\emph{Norm-based}}.\footnote{Note that in most of our experiments, we use all these methods within the rollout aggregation technique.} 
The \emph{Weight-based} approaches employed in our experiments are as follows:
\begin{itemize}
    \itemsep0em 
    \item $\mathcal{W}$\,:
        The raw attention maps averaged across all heads (See $\overline{\boldsymbol{A}}_{\ell}$ in \S\ref{sec:background}).

    \item $\mathcal{W_\textsc{FixedRes}}$\,:
        \citeauthor{abnar-zuidema-2020-quantifying}'s assumption; add an identity matrix as a fixed residual to $\overline{\boldsymbol{A}}_{\ell}$ (see $\hat{\boldsymbol{A}}_{\ell}$ in Eq. \ref{eq:attention_rollout_residual}).
    
    \item $\mathcal{W_\textsc{Res}}$\,: 
        The corrected version of $\mathcal{W}$ in which accurate residuals are added based on the context-mixing ratios of $\mathcal{N_\textsc{Enc}}$:
        
        \begin{equation}
        \label{eq:r_nenc}
        \begin{aligned}
        \hat{r}_{i}=&\frac{\left\|\sum_{j=1, j \neq i}^{n} \tilde{\boldsymbol{x}}_{i \leftarrow j}\right\|}{\left\|\sum_{j=1, j \neq i}^{n} \tilde{\boldsymbol{x}}_{i \leftarrow j}\right\|+\left\|\tilde{\boldsymbol{x}}_{i \leftarrow i}\right\|} \\
        \end{aligned}
        \end{equation}
        In order to enforce $\mathcal{W_\textsc{Res}}$ to have a context-mixing ratio equal to $\hat{r}_{i}$, it is essential to \mbox{zero-out} the diagonal elements (the tokens' attentions to themselves) of $\bar{\bm{A}}_{\ell}$ and renormalize it:
        \begin{equation}
        \label{eq:w_res}
        \begin{gathered}
        \bm{A}_{\ell}^{\prime}=
        (\bm{I} -  \operatorname{\textbf{diag}}\left(\bar{\bm{A}}_{\ell}\right))^{-1}(\bar{\bm{A}}_{\ell} - 
        \operatorname{\textbf{diag}}\left(\bar{\bm{A}}_{\ell}\right)) \\
        \begin{aligned}
        \mathcal{W_\textsc{Res}}:=&
        \operatorname{\textbf{diag}}\left(\hat{r}_{1},\cdots,\hat{r}_{n}\right)\bm{A}_{\ell}^{\prime} \\ +&\operatorname{\textbf{diag}}\left(1-\hat{r}_1, \ldots, 1-\hat{r}_{n}\right)\bm{I}
        \end{aligned}
        \end{gathered}
        \end{equation}
    
\end{itemize}
The \emph{Norm-based} analysis methods, namely $\mathcal{N}$, $\mathcal{N_\textsc{Res}}$ and $\mathcal{N_\textsc{ResLN}}$ were discussed in detail in \S\ref{sec:background}. Our proposed norm-based method $\mathcal{N_\textsc{Enc}}$ was explained in \S\ref{sec:methodology}. 
For an ablation study, we introduce $\mathcal{N_\textsc{FixedRes}}$ which is $\mathcal{N}$, corrected with a fixed residual similar to $\mathcal{W_\textsc{FixedRes}}$\footnote{The only difference is that we need to normalize $\mathcal{N}$ before adding an identity matrix.}.

\begin{equation}
\begin{gathered}
    \hat{\mathcal{N}} = \left (\frac{||\bm{z}_{i\leftarrow j}||}{\sum_j||\bm{z}_{i\leftarrow j}||}\right) \in \mathbb{R}^{n\times n}\\
    \mathcal{N_\textsc{FixedRes}} := 0.5\,\hat{\mathcal{N}}+0.5\,\bm{I}
\end{gathered}
\end{equation}

In \S\ref{sec:results}, we will demonstrate our comparative studies between the aforementioned methods and \methodName.

\subsection{Gradient-based Methods for Faithfulness Analysis}
Gradient-based methods are widely used as alternatives for attention-based counterparts for quantifying the importance of a specific input feature in making the right prediction \cite{li-etal-2016-visualizing, atanasova-etal-2020-diagnostic}.
In this section we discuss the specific gradient-based methods we use, namely saliency, HTA, and our adjusted HTA.

\subsubsection{Saliency}
Gradient-based saliency is based on the gradient of the output ($y_{c}$) w.r.t. the input embeddings ($\boldsymbol{e}_{i}^{0}$). One of the most accurate variations of the saliency family is the \emph{gradient$\times$input} method \citep{kindermans2016investigating} where the input embeddings is multiplied by the gradients. Thus, the contribution score of input token $i$ is determined by first computing the element-wise product of the input embeddings ($\boldsymbol{e}_{i}^{0}$) and the gradients of the true class output score ($y_{c}$) w.r.t. the input embeddings. Then, the L2 norm of the scaled gradients is computed to derive the final score:
\begin{equation}
    Saliency_{i}=\left\|\frac{\partial y_{c}}{\partial \boldsymbol{e}_{i}^{0}} \odot \boldsymbol{e}_{i}^{0}\right\|_{2}
\end{equation}

\subsubsection{HTA x Inputs}
\label{sec:HTA_x_inputs}
To determine an upper bound on the information mixing within each layer, we use a modified version of \emph{Hidden Token Attribution} \citep[HTA]{Brunner2020On}. In the original version, HTA is the sensitivity between any two vectors in the model's computational graph. However, inspired by the \emph{gradient$\times$input} method \citep{kindermans2016investigating}, which has shown more faithful results \citep{atanasova-etal-2020-diagnostic, on_explaining_your_explanationsBERT}, we multiply the input vectors by the gradients and then apply a Frobenius norm. We compute the attribution from hidden embedding $j$ ($\bm e^{\ell-1}_j$) to hidden embedding $i$ ($\bm e^{\ell}_i$) in layer $\ell$ as:
\begin{equation}
\label{eq:HTA_x_Input}
\bm{c}_{i\leftarrow j}^\ell = \left\|\frac{\partial \bm e^{\ell}_i}{\partial \bm e^{\ell-1}_j} \odot \bm e^{\ell-1}_j\right\|_F
\end{equation}
Computing HTA-based attribution matrices is an extremely computation-intensive task (especially for long texts) due to the high dimensionality of hidden embeddings. 
Hence, we only use this method for 256 examples from the SST-2 task's validation set. It is worth noting that extracting the HTA-based contribution maps for the aforementioned data took approximately 2 hours, whereas computing the maps for the entire analysis methods stated in \S\ref{sec:analysis_methods} took only 5 seconds.\footnote{Conducted on a 3070 GPU machine.}

\subsection{Setup}
We employ HuggingFace's Transformers library\footnote{\href{https://github.com/huggingface/transformers}{https://github.com/huggingface/transformers}} \cite{wolf-etal-2020-transformers} and the BERT-base-uncased model. 
For fine-tuning BERT, epochs vary from 3 to 5, and the batch size and learning rate are 32 and 3e-5, respectively.\footnote{Recommended by \citet{devlin-etal-2019-bert}.}
We also carried out the main experiment on BERT-large and ELECTRA \cite{devlin-etal-2019-bert, clark2020electra} where the results are reported at \S\ref{more_models}.

After rollout aggregation of each analysis method, we obtain an accumulated attribution matrix for every layer ($\ell$) of BERT. These matrices indicate the overall contribution of each input token to all token representations in layer $\ell$.
Since the classifier in a fine-tuned model is attached to the final layer representation of the \textsc{[CLS]} token, we consider the first row (corresponding to \textsc{[CLS]} attributions) of the last layer attribution matrix.
This vector represents the contribution of each input token to the model's final decision.
As a measure of faithfulness of the resulting vector with the saliency scores, we report the \emph{Spearman’s rank correlation} between the two vectors.

\subsection{Results}
\begin{table*}[t!]
\begin{center}
% \small
\tabcolsep=0.13cm
\begin{tabular}{l c c c} 
 \toprule
      & \multicolumn{3}{c}{\textbf{Attention Rollout}} \\
    \cmidrule(lr){2-4}
      & 
     \textsc{SST2} & 
     \textsc{MNLI} & 
     \textsc{HateXplain} \\
    % \textbf{Task} & F1 Score & Compression & F1 Score & Compression & F1 Score & Compression \\
    \midrule
    Weight-based \small{$(\mathcal{W})$}
                                    & \textminus 0.11 ± 0.26      & \textminus 0.06 ± 0.22 & \enspace0.12 ± 0.26 \\
    \qquad w/ Fixed Residual \small{$(\mathcal{W_\textsc{FixedRes}})$} \footnotemark
                                    & \textminus 0.24 ± 0.26      & \textminus 0.05 ± 0.26 & \enspace0.13 ± 0.28 \\
    \qquad w/ Residual \small{$(\mathcal{W_\textsc{Res}})$}
                                    &  \enspace0.19 ± 0.26      &  \enspace0.27 ± 0.25 &  \enspace0.53 ± 0.24 \\
    \midrule
    Norm-based \small{$(\mathcal{N})$} %\small{$^\blacklozenge$}
                                    & \enspace0.44 ± 0.20      & \enspace0.47 ± 0.16 & \enspace0.43 ± 0.22 \\
    \qquad w/ Fixed Residual \small{$(\mathcal{N_\textsc{FixedRes}})$}
                                    & \enspace0.48 ± 0.20      & \enspace0.55 ± 0.16 & \enspace0.48 ± 0.22 \\
    \qquad w/ Residual \small{$(\mathcal{N_\textsc{Res}})$} %\small{$^\spadesuit$}
                                    & \enspace0.73 ± 0.13 & \enspace0.75 ± 0.10 & \enspace0.66 ± 0.17 \\
    \qquad w/ Residual + Layer Norm 1 \small{$(\mathcal{N_\textsc{ResLN}})$} %\small{$^\spadesuit$}
                                    & \textminus 0.21 ± 0.26  & \textminus 0.06 ± 0.26 & \enspace0.08 ± 0.28 \\
    \qquad w/ \textbf{\methodName}: [Residual + Layer Norm 1, 2] \small{$(\mathcal{N_\textsc{Enc}})$} %\small{$^\spadesuit$}
                                &  \bf{\enspace0.77 ± 0.12}  & \bf{\enspace0.78 ± 0.09} & \bf{\enspace0.72 ± 0.17} \\
 \bottomrule
\end{tabular}
\end{center}
\captionsetup{aboveskip=0pt}
\caption{
% \red{(replace + with space "~~". As of now, it looks like a \textit{relative} improvement with all these plus and minus signs.)}  # Done
Spearman's rank correlation of attribution based importance (aggregated by rollout) with saliency scores for the validation set for the BERT model fine-tuned on SST-2, MNLI, and HateXplain. 
In fixed residual cases, the context-mixing ratio is roughly $0.5$, and in weight-based w/ residual $(\mathcal{W_\textsc{Res}})$, it is corrected with context-mixing ratio of $(\mathcal{N_\textsc{Enc}})$. The numbers are the average on all the validation set examples ± the standard deviation.
}

\label{tab:spearman}
\end{table*}

\label{sec:results}

Table~\ref{tab:spearman} shows the Spearman correlation of saliency scores with the aggregated attribution scores from \textsc{[CLS]} to input tokens at the final layer. 
In order to determine the contribution of each component of encoder layer to the overall performance, we report the results for attribution analysis methods discussed in \S\ref{sec:analysis_methods}.
Our results demonstrate that incorporating the vector norms, residual connection, and both layer normalizations yields the highest correlation ($\mathcal{N_\textsc{Enc}}$). 
In what follows, we discuss the impact of incorporating various parts in the analysis.

\begin{figure}[t]
\centering
    \includegraphics[width=0.49\textwidth]{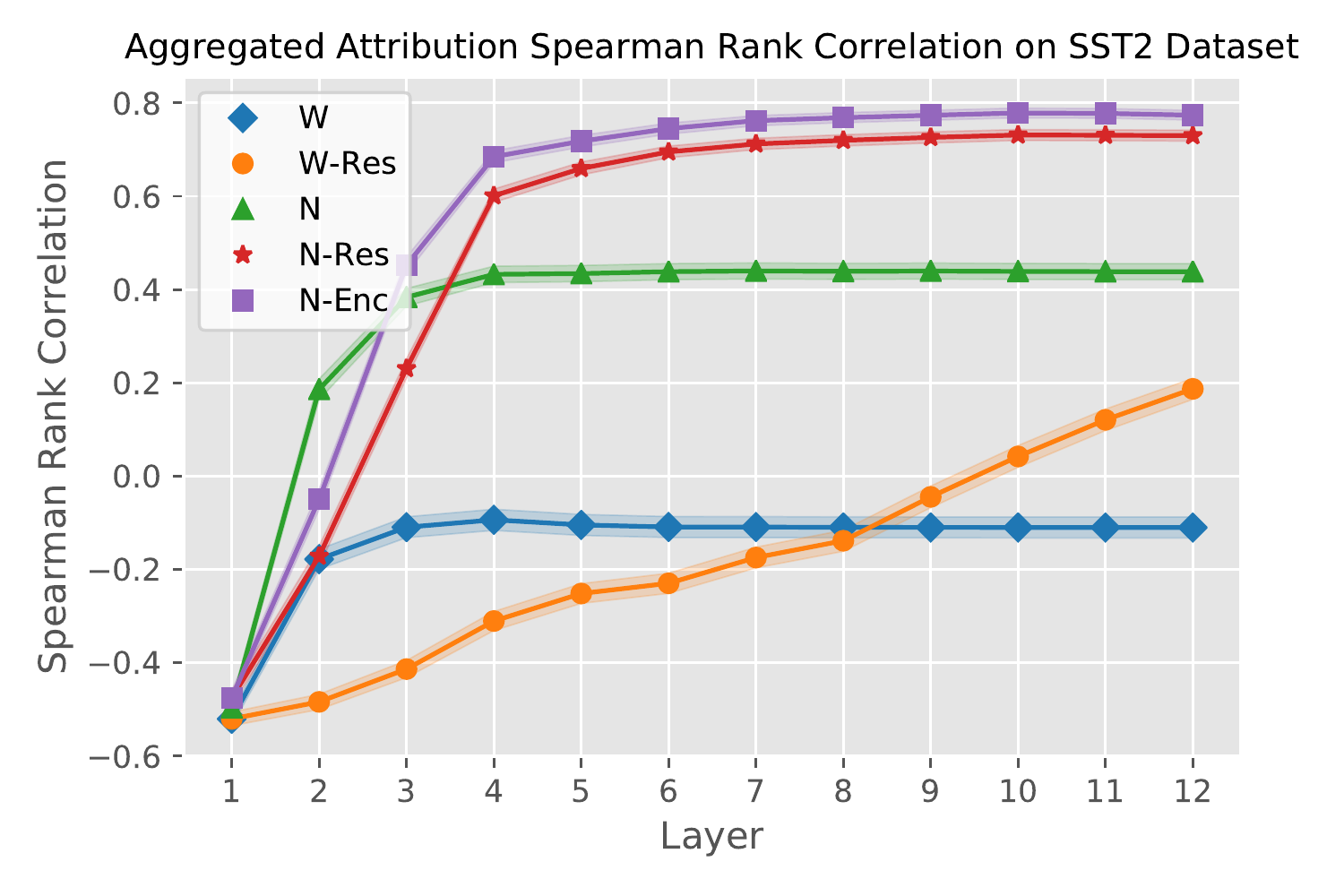}
    \caption{Spearman's rank correlation of aggregated attribution scores with saliency scores across layers. The 99\% confidence intervals are shown as (narrow) shaded areas around each line. $\mathcal{N_\textsc{Enc}}$ achieves the highest correlation in almost every layer.}
    \label{fig:sst2_spearman}
\end{figure}

\subsubsection{On the role of vector norms}
As also suggested by \citet{kobayashi-etal-2020-attention-norm}, vector norms play an important role in determining attention outputs. 
This is highlighted by the significant gap between weight-based and norm-based settings across all datasets in Table~\ref{tab:spearman}. 

We also show the correlation of the aggregated attention for all layers in Figure~\ref{fig:sst2_spearman}. 
The norm-based settings ($\mathcal{N}$ and $\mathcal{N_\textsc{Res}}$) attain higher correlation than the weight-based counterparts ($\mathcal{W}$ and $\mathcal{W_\textsc{Res}}$) almost in all layers, confirming the importance of incorporating vector norms.

\footnotetext{
As mentioned in \S4.2, this analysis method is based on the original experiment by \citet{abnar-zuidema-2020-quantifying}. Our experiments on SST2 differ from theirs in two aspects: (i) we opted for \emph{gradient$\times$input} saliencies, while they used the sum of gradients (sensitivity) (ii) instead of BERT, they used a DistillBERT fine-tuned model \citep{sanh2019distilbert}. 
However, their reported results in their sepcific setup (Spearman Corr. = 0.14) still yields significantly lower results than \methodName.
}
 
\subsubsection{On the role of residual connections}
\citet{kobayashi-etal-2021-incorporating-residual} showed that in the encoder layer, the output representations of each token is mainly determined by its own representation, and the contextualization from other tokens' plays a marginal role.
This is in contrary to the simplifying assumption made by \citet{abnar-zuidema-2020-quantifying} who used a fixed context-mixing ratio of $0.5$ (assuming that BERT equally preserves and mixes the representations).
This setting is shown as weight-based with fixed residual ($\mathcal{W_\textsc{FixedRes}}$) in Table~\ref{tab:spearman}.
We compare this setting against $\mathcal{W_\textsc{Res}}$ (see \S\ref{sec:analysis_methods}). 
$\mathcal{W_\textsc{Res}}$ is similar to $\mathcal{W_\textsc{FixedRes}}$ (in that it does not take into account vector norms) but differs in that it considers a dynamic mixing ratio (the one from $\mathcal{N_\textsc{Enc}}$).
The huge performance gap between the two settings in Table~\ref{tab:spearman} clearly highlights the importance of considering accurate context-mixing ratios.
Therefore, it is crucial to consider the residual connection in the attention block for input token attribution analysis.

To further demonstrate the role of residual connections, we utilize the introduced method in \S\ref{sec:analysis_methods}, where we modified the norm-based attentions with fixed residual ($r\approx0.5$).
The comparison of norm-based without any residual ($\mathcal{N}$) and with a fixed residual ($\mathcal{N_\textsc{FixedRes}}$) shows a consistent improvement for the latter across all the datasets. 
This provides evidence on that having a fixed uniform context-mixing ratio is better than neglecting the residual connection altogether.

Finally, when we aggregate the norm-based analysis with an accurate dynamic context-mixing ratio ($\mathcal{N_\textsc{Res}}$), we observe the highest correlation up to this point, without layer normalization.

\subsubsection{On the role of layer normalization}
\begin{figure}[t]
\centering
    \includegraphics[width=0.49\textwidth]{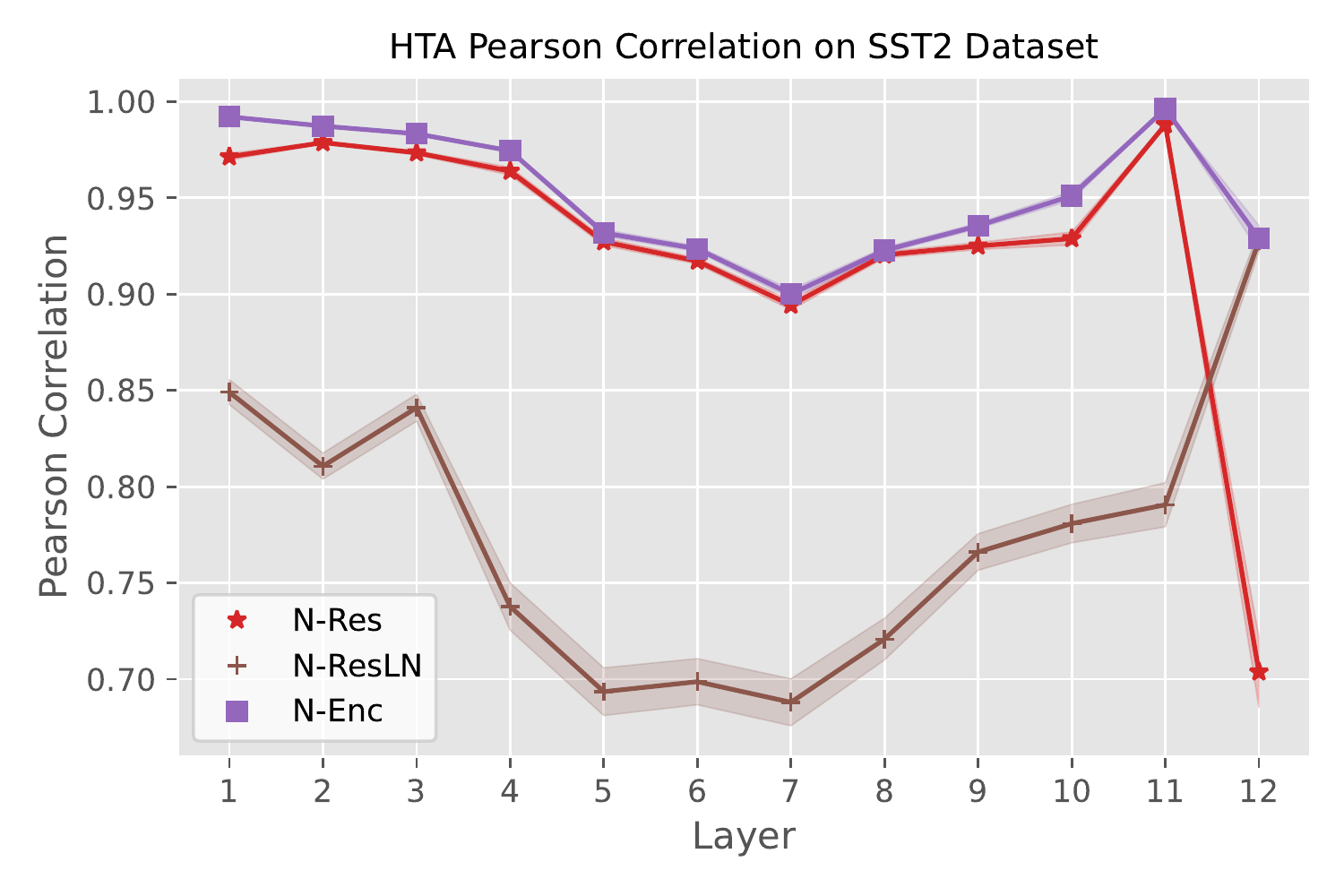}
    \caption{Single layer Pearson correlation of HTA maps with attribution maps. The 99\% confidence intervals are shown as shaded areas around each line. $\mathcal{N_\textsc{ResLN}}$ shows considerably less association with HTA.
    }
    \label{fig:sst2_hta_corr}
\end{figure}
In Table~\ref{tab:spearman} we see a sudden drop in correlations for $\mathcal{N_\textsc{ResLN}}$. 
Although this method considers vector norms and residuals, incorporating LN\#1 in the encoder seems to have deteriorated the accuracy for token attribution analysis.
To determine whether this deterioration of correlation in aggregated attributions is also present in individual single layers, we compare the HTA maps as a baseline with the attribution matrices extracted from different analysis methods.
Figure~\ref{fig:sst2_hta_corr} shows the correlation of HTA attribution maps with the maps obtained by $\mathcal{N_\textsc{Res}}$, $\mathcal{N_\textsc{ResLN}}$, and $\mathcal{N_\textsc{Enc}}$ methods. 
The results indicate that $\mathcal{N_\textsc{ResLN}}$ exhibits a significantly lower association. 

The question that arises here is that how incorporating an additional component of the encoder (LN\#1) in $\mathcal{N_\textsc{ResLN}}$ degrades the results (compared to $\mathcal{N_\textsc{Res}}$).
\begin{figure}[t]
\centering
    \includegraphics[width=0.49\textwidth]{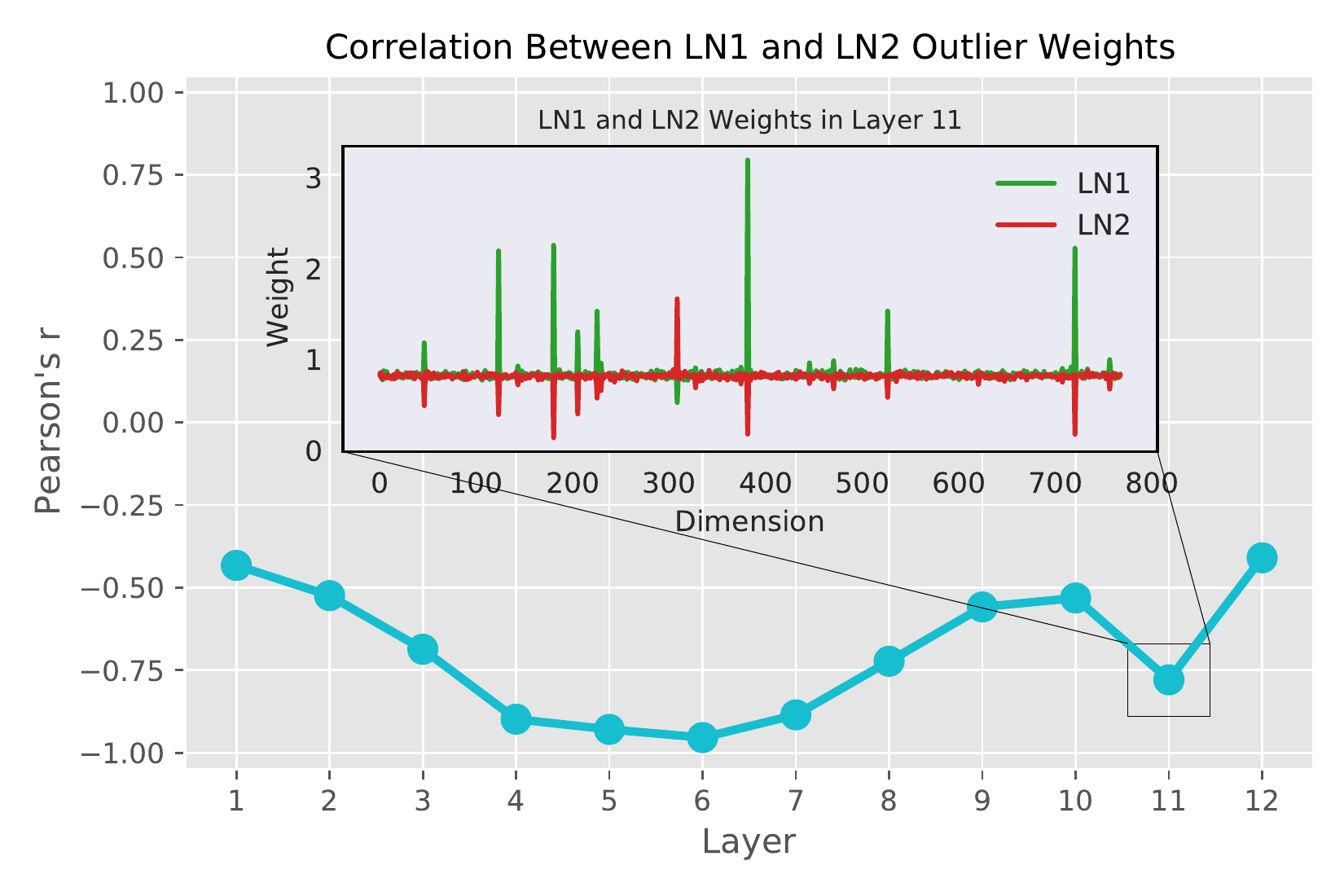}
    \caption{The Pearson correlation between outlier weights of LN\#1 and LN\#2 across layers. The weight values for layer 11 are shown as well.
    }
    \label{fig:ln_weights}
\end{figure}
To answer this question, we investigated the learned weights of LN\#1 and LN\#2.
The outlier weights\footnote{We identify the dimensions where the weights are at least $3\sigma$ from the mean as outliers \cite{kovaleva-etal-2021-bert}.}
in specific dimensions of LNs are shown to be significantly influential on the model's performance \cite{kovaleva-etal-2021-bert, luo-etal-2021-positional}.
It is interesting to note that based on our observations, the outlier weights of the two layer norms seem to be the opposite of each other. 
Figure~\ref{fig:ln_weights} demonstrates the weight values in layer 11 and also the correlation of the outlier weights across layers. The large negative correlations confirm that the outlier weights work contrary to each other.
We speculate that the effect of outliers in the two layer norms is partly cancelled out when both are considered.
 
As shown in Figure~\ref{fig:encoder_diagram}, the FFN and the second layer normalization are on top of the attention block. However, $\mathcal{N_\textsc{ResLN}}$ does not incorporate the components outside of the attention block.
As described in \S\ref{sec:methodology}, in our local analysis method $\mathcal{N_\textsc{Enc}}$ we incorporate the second layer normalization in the transformer's encoder (Figure~\ref{fig:encoder_diagram}), thus considering the whole encoder block (except FFN). 
Overall, our global method, \methodName, yields the best results among all the methods evaluated in our experiments. 
In general, Table~\ref{tab:spearman} suggests that incorporating each component of the encoder will increase the correlation; however, the two layer normalizations should be considered together.

\begin{table}[t!]
\begin{center}
\small
\tabcolsep=0.11cm
\begin{tabular}{c l | c | c | c | c} 
    \toprule
    & & L1 & L6 & L12 & MAX \\
    \midrule
    \multirow{3}{*}{\rotatebox[origin=c]{90}{Indiv.}}
    & $\mathcal{N}$ & 
        \textminus .50 ± .18 & +.28 ± .23 & +.40 ± .21 & +.41 ± .21 \\
    & $\mathcal{N_\textsc{Res}}$ & 
        \textminus .48 ± .18 & +.29 ± .24 & +.41 ± .19 & +.41 ± .19 \\
    & $\mathcal{N_\textsc{Enc}}$ & 
        \bf{\textminus .47 ± .18} & +.29 ± .24 & +.41 ± .19 & +.41 ± .19 \\
    \midrule
    \multirow{3}{*}{\rotatebox[origin=c]{90}{Rollout}}
    & $\mathcal{N}$ & 
        \textminus .50 ± .18 & +.44 ± .20 & +.44 ± .20 & +.44 ± .20 \\
    & $\mathcal{N_\textsc{Res}}$ & 
        \textminus .48 ± .18 & +.70 ± .14 & +.73 ± .13 & +.73 ± .13 \\ 
    & $\mathcal{N_\textsc{Enc}}$ & 
        \bf{\textminus .47 ± .18} & \bf{+.74 ± .14} & \bf{+.77 ± .12} & \bf{+.78 ± .12} \\   
    \bottomrule
\end{tabular}
\end{center}
\captionsetup{aboveskip=0pt}
\caption{
Spearman's rank correlation of attribution-based scores (individual and aggregated by rollout) with saliency scores for the validation set for the BERT model fine-tuned on SST-2. The results are reported for layers 1, 6, 12, and the maximum of all layers.
Utilizing rollout aggregation achieves higher correlations than individual layers.
}

\label{tab:aggregation}
\end{table}

\subsubsection{On the role of aggregation}

We carried out an additional analysis to verify if incorporating vector norms, residual connection and layer normalizations in individual layers is adequate for achieving high correlations, or if it is also necessary to aggregate them via rollout. 
Table~\ref{tab:aggregation} shows the correlation results in different layers for raw attributions (without aggregation) and for the aggregated attributions using the rollout method.
Applying rollout method on attribution maps up to each layer results in higher correlations with the saliency scores than the raw single layer attribution maps, especially in deeper layers.
Therefore, attention aggregation is essential for global input token attribution analysis. 

An interesting point in Figure~\ref{fig:sst2_spearman}, which shows the correlation of the aggregated methods throughout the layers, is that the correlation curves flatten out after only a few layers.\footnote{
$\mathcal{W_\textsc{Res}}$ is the only exception with a constant increase; this method is gradually and artificially corrected by $\mathcal{N_\textsc{Enc}}$ context mixing ratios.
}
This indicates that BERT identifies decisive tokens only after the first few layers.
The final layers only make minor adjustments to this order. Nevertheless, it is worth noting that the order of attribution does not necessarily imply the model's final decision and the final result may still change for the better or worse \cite{zhou2020bert}.

\subsubsection{Qualitative analysis}
To qualitatively answer if the aggregated attribution maps provide plausible and meaningful interpretations, we take a closer look at the attribution maps generated by \methodName. 
Figure~\ref{fig:sst2_qualitative} shows the \methodName~attribution of the model trained on SST-2. 
Each layer demonstrates the \textsc{[CLS]} token's aggregated attribution to input tokens up to the corresponding layer. 
The example inputs are 
``a deep and meaningful film.'' and ``big fat waste of time.'', both correctly classified by the model.
In both cases, \methodName~focuses on the relevant words for sentiment classification, i.e., ``meaningful'' and ``waste''.
An interesting observation in Figure~\ref{fig:sst2_qualitative} is that in the first few layers, the \textsc{[CLS]} token mostly attends to itself while other tokens have marginal impact.
As the representations get more contextualized in deeper layers, the attribution correctly shifts to the words which indicate the sentiment of the sentence.\footnote{Complete attention maps in Figure~\ref{fig:sst2_tokens} show that, similarly to \textsc{[CLS]}, other tokens also focus on sentiment tokens.}
More examples from MNLI and SST2 datasets, including misclassified examples are available at \S\ref{more_examples}.
Our qualitative analysis suggests that \methodName~can be useful for a reasonable interpretation of attention mechanism in BERT, ELECTRA, and possibly any other transformer-based model.

\section{Related Work}
While numerous studies have used attention weights to analyze and interpret the self-attention mechanism \citep{clark-etal-2019-bert-look, kovaleva-etal-2019-revealing, reif2019visualizing, bert-track-syntactic-dep}, the use of mere attention weights to explain a model's inner workings has been an active topic of debate \citep{serrano-smith-2019-attention, jain-wallace-2019-attention, wiegreffe-pinter-2019-attention}. 
Several solutions have been proposed to address this issue, usually through converting raw attention weights to scores that provide better explanations.
\citet{Brunner2020On} used the transformation function $f^h(\bm{x}_j)$ to introduce \emph{effective attentions}\textemdash the orthogonal component of the attention matrix in $f^h(\bm{x}_j)$ null space\textemdash to explain the inner workings of each layer. 
However, this technique ignores other components in the encoder and is computationally expensive due to the SVD required to compute the effective attentions. 
\citet{kobayashi-etal-2020-attention-norm} incorporated the modified vector and introduced a vector norms-based analysis. This was later extended by integrating residual connections and layer normalization components to enhance the accuracy of explanations \cite{kobayashi-etal-2021-incorporating-residual}. 
But, as discussed in \S\ref{sec:results}, relying solely on LN\#1
does not produce accurate results.

While these methods can be employed for single-layer (local) analysis, multi-layer attributions are not necessarily correlated with single-layer attributions due to the significant degree of information combination through multi-layer language models \citep{pascual-etal-2021-telling-full-story, Brunner2020On}.
Various saliency methods exist for explaining the model's decision based on the input \citep{li-etal-2016-visualizing, bastings-filippova-2020-elephant, atanasova-etal-2020-diagnostic, on_explaining_your_explanationsBERT, mohebbi-etal-2021-exploring}. However, these approaches are not primarily designed for computing inter-token attributions.
To fill this gap, \citet{Brunner2020On} proposed HTA, which is based on the gradient of each hidden embedding in relation to the input embeddings.
In \S\ref{sec:HTA_x_inputs}, we extend HTA to incorporate the impact of the input vectors. 
However, HTA is extremely computationally intensive. 
Attention rollout (see \S\ref{sec:methodology}) and attention flow\textemdash which involve solving a max-flow problem on the attention graph\textemdash are two aggregation approaches introduced by \citet{abnar-zuidema-2020-quantifying}, in which raw attention weights (with equally weighted residual weights) are aggregated within multiple layers. 
We showed that attention rollout does not perform well on the raw attention maps of language models fine-tuned on downstream tasks and that this problem can be resolved by utilizing attribution norms.

\section{Conclusions}
In this work, we proposed a novel method for single layer token attribution analysis which incorporates the whole encoder layer, i.e., the attention block and the output layer normalization. 
When aggregated across layers using the rollout method, our technique achieves quantitatively and qualitatively plausible results.
Our evaluation of different analysis methods provided evidence on roles played by individual components of the encoder layer, i.e., the vector norms, the residual connections, and the layer normalizations.
Furthermore, our in-depth analysis suggested that the two layer normalizations in the encoder layer counteract each other; hence, it is important to couple them for an accurate analysis.

Additionally, using a newly proposed and improved version of Hidden Token Attribution, we demonstrated that encoder-based attribution analysis is more accurate when compared to other partial solutions in a single layer (local-level). This is consistent with our global observations.
Quantifying global input token attribution based on our work can provide a meaningful explanation of the whole model's behavior. In future work, we plan to apply our global analysis method on various datasets and models, to provide valuable insights into model decisions and interpretability.

% \section*{Acknowledgements}
% ...

% \newpage

\bibliography{anthology,custom}
\bibliographystyle{acl_natbib}

\appendix
\counterwithin{figure}{section}
\counterwithin{table}{section}
\section{Appendix}
\label{sec:appendix}

\subsection{LN Formulation}
\label{sec:formulation}
$m(\bm a):=\frac{1}{d}\sum_{k}\bm{a}^{(k)}$, \\\\
$s(\bm a):=\sqrt{\frac{1}{d}\sum_{k}(m(\bm{a}) - \bm{a}^{(k)} + \epsilon)^2}$ 
\\ where $\epsilon$ is a small constant

\subsection{More Models}
\label{more_models}
In this section we provide the results for BERT-large and ELECTRA-base. For both models, our method outperforms the previous analysis methods. The results are reported in Tables \ref{tab:spearman_bert_large} and \ref{tab:spearman_electra}.
\begin{table*}[t!]
\begin{center}
% \small
\tabcolsep=0.13cm
\begin{tabular}{l c c c} 
 \toprule
      BERT-large & \multicolumn{3}{c}{\textbf{Attention Rollout}} \\
    \cmidrule(lr){2-4}
      & 
     \textsc{SST2} & 
     \textsc{MNLI} & 
     \textsc{HateXplain} \\
    \midrule
    Weight-based \small{$(\mathcal{W})$}
                                    & \textminus 0.38 ± 0.16      & \textminus 0.61 ± 0.14 & \textminus 0.41 ± 0.25 \\
    \qquad w/ Fixed Residual \small{$(\mathcal{W_\textsc{FixedRes}})$}
                                    & \textminus 0.25 ± 0.19      & \textminus 0.48 ± 0.19 & \textminus 0.21 ± 0.30 \\
    \qquad w/ Residual \small{$(\mathcal{W_\textsc{Res}})$}
                                    & \textminus 0.10 ± 0.21      &  \enspace 0.33 ± 0.23 &  \enspace 0.09 ± 0.30 \\
    \midrule
    Norm-based \small{$(\mathcal{N})$} %\small{$^\blacklozenge$}
                                    & \enspace0.44 ± 0.24      & \enspace 0.13 ± 0.27 & \enspace 0.48 ± 0.25 \\
    \qquad w/ Fixed Residual \small{$(\mathcal{N_\textsc{FixedRes}})$}
                                    & \enspace0.49 ± 0.24      & \enspace 0.26 ± 0.25 & \enspace 0.49 ± 0.30 \\
    \qquad w/ Residual \small{$(\mathcal{N_\textsc{Res}})$} %\small{$^\spadesuit$}
                                    & \enspace0.77 ± 0.11 & \enspace 0.66 ± 0.12 & \enspace 0.73 ± 0.16 \\
    \qquad w/ Residual + Layer Norm 1 \small{$(\mathcal{N_\textsc{ResLN}})$} %\small{$^\spadesuit$}
                                    & \textminus 0.07 ± 0.23  & \textminus 0.35 ± 0.24 & \enspace 0.06 ± 0.32 \\
    \qquad w/ \textbf{\methodName}: [Residual + Layer Norm 1, 2] \small{$(\mathcal{N_\textsc{Enc}})$} %\small{$^\spadesuit$}
                                &  \bf{\enspace0.83 ± 0.08}  & \bf{\enspace 0.77 ± 0.09} & \bf{\enspace 0.76 ± 0.17} \\
 \bottomrule
\end{tabular}
\end{center}
\captionsetup{aboveskip=0pt}
\caption{
Spearman's rank correlation of attribution based importance (aggregated by rollout) with saliency scores for the validation set for the BERT-large model fine-tuned on SST-2, MNLI, and HateXplain. The numbers are the average on all the validation set examples (1024 examples for MNLI dataset due to resource limitations) ± the standard deviation.
}

\label{tab:spearman_bert_large}
\end{table*}
\begin{table*}[t!]
\begin{center}
% \small
\tabcolsep=0.13cm
\begin{tabular}{l c c c} 
 \toprule
      ELECTRA-base & \multicolumn{3}{c}{\textbf{Attention Rollout}} \\
    \cmidrule(lr){2-4}
      & 
     \textsc{SST2} & 
     \textsc{MNLI} & 
     \textsc{HateXplain} \\
    % \textbf{Task} & F1 Score & Compression & F1 Score & Compression & F1 Score & Compression \\
    \midrule
    Weight-based \small{$(\mathcal{W})$}
                                    & \textminus 0.37 ± 0.19      & \textminus 0.31 ± 0.22 & \enspace0.02 ± 0.29 \\
    \qquad w/ Fixed Residual \small{$(\mathcal{W_\textsc{FixedRes}})$}
                                    & \textminus 0.37 ± 0.19      & \textminus 0.24 ± 0.23 & \enspace0.01 ± 0.29 \\
    \qquad w/ Residual \small{$(\mathcal{W_\textsc{Res}})$}
                                    & \textminus 0.10 ± 0.22      &  \enspace0.08 ± 0.25 &  \enspace0.20 ± 0.27 \\
    \midrule
    Norm-based \small{$(\mathcal{N})$} %\small{$^\blacklozenge$}
                                    & \enspace0.18 ± 0.21      & \enspace0.12 ± 0.21 & \enspace0.21 ± 0.26 \\
    \qquad w/ Fixed Residual \small{$(\mathcal{N_\textsc{FixedRes}})$}
                                    & \enspace0.23 ± 0.22      & \enspace0.32 ± 0.23 & \enspace0.28 ± 0.26 \\
    \qquad w/ Residual \small{$(\mathcal{N_\textsc{Res}})$} %\small{$^\spadesuit$}
                                    & \enspace0.54 ± 0.17 & \enspace0.54 ± 0.14 & \enspace0.44 ± 0.21 \\
    \qquad w/ Residual + Layer Norm 1 \small{$(\mathcal{N_\textsc{ResLN}})$} %\small{$^\spadesuit$}
                                    & \textminus 0.24 ± 0.23  & \textminus 0.16 ± 0.24 & \textminus 0.07 ± 0.28 \\
    \qquad w/ \textbf{\methodName}: [Residual + Layer Norm 1, 2] \small{$(\mathcal{N_\textsc{Enc}})$} %\small{$^\spadesuit$}
                                &  \bf{\enspace0.64 ± 0.15}  & \bf{\enspace0.68 ± 0.12} & \bf{\enspace0.47 ± 0.22} \\
 \bottomrule
\end{tabular}
\end{center}
\captionsetup{aboveskip=0pt}
\caption{
% \red{(replace + with space "~~". As of now, it looks like a \textit{relative} improvement with all these plus and minus signs.)}  # Done
Spearman's rank correlation of attribution based importance (aggregated by rollout) with saliency scores for the validation set for the ELECTRA-base model fine-tuned on SST-2, MNLI, and HateXplain. The numbers are the average on all the validation set examples ± the standard deviation.
}

\label{tab:spearman_electra}
\end{table*}
\begin{figure*}[t]
\centering
    \subfloat{
        % left bottom right top
        \includegraphics[width=0.23\textwidth, trim=35 15 50 50, clip] {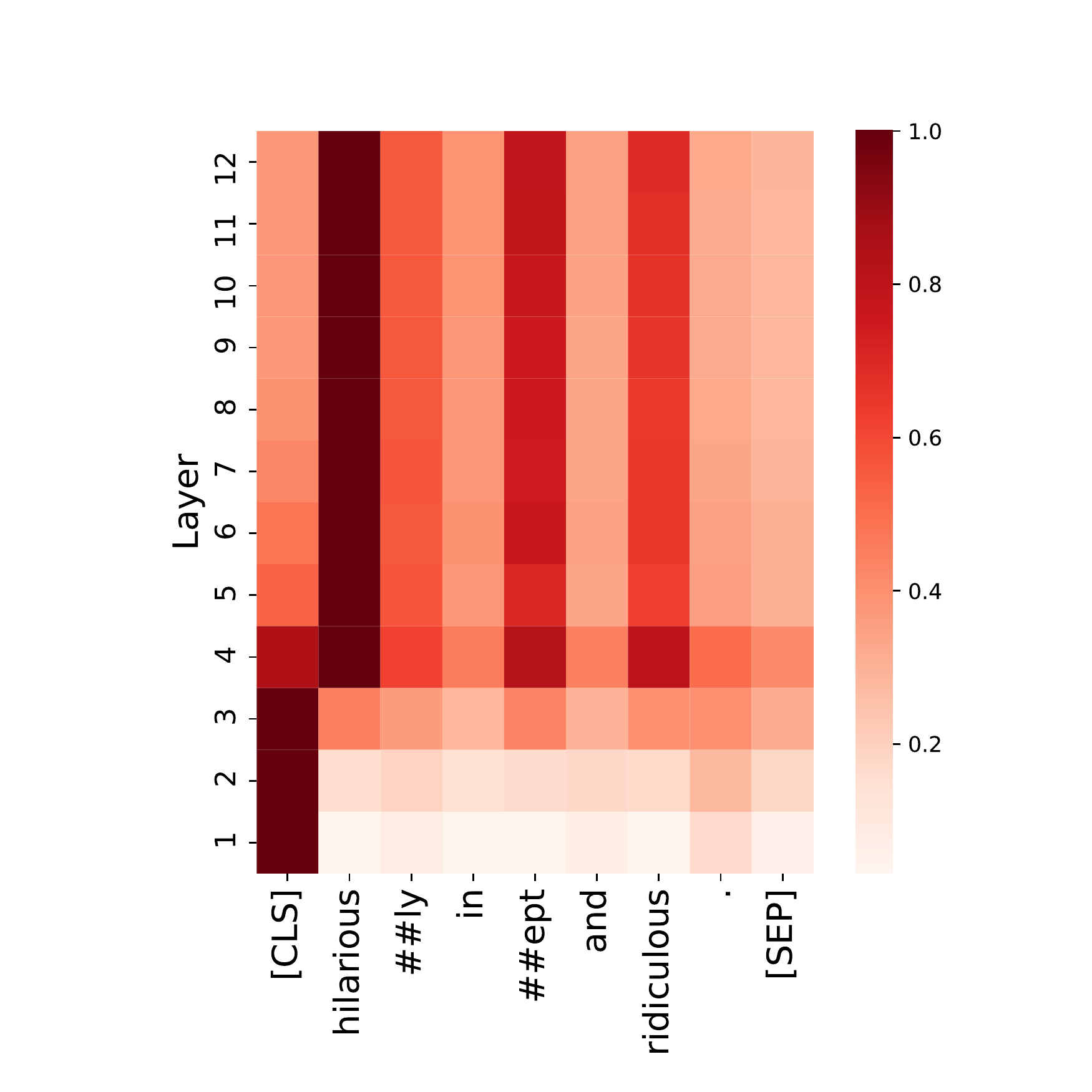}
    }
    \subfloat{
        \includegraphics[width=0.25\textwidth, trim=25 15 50 50, clip] {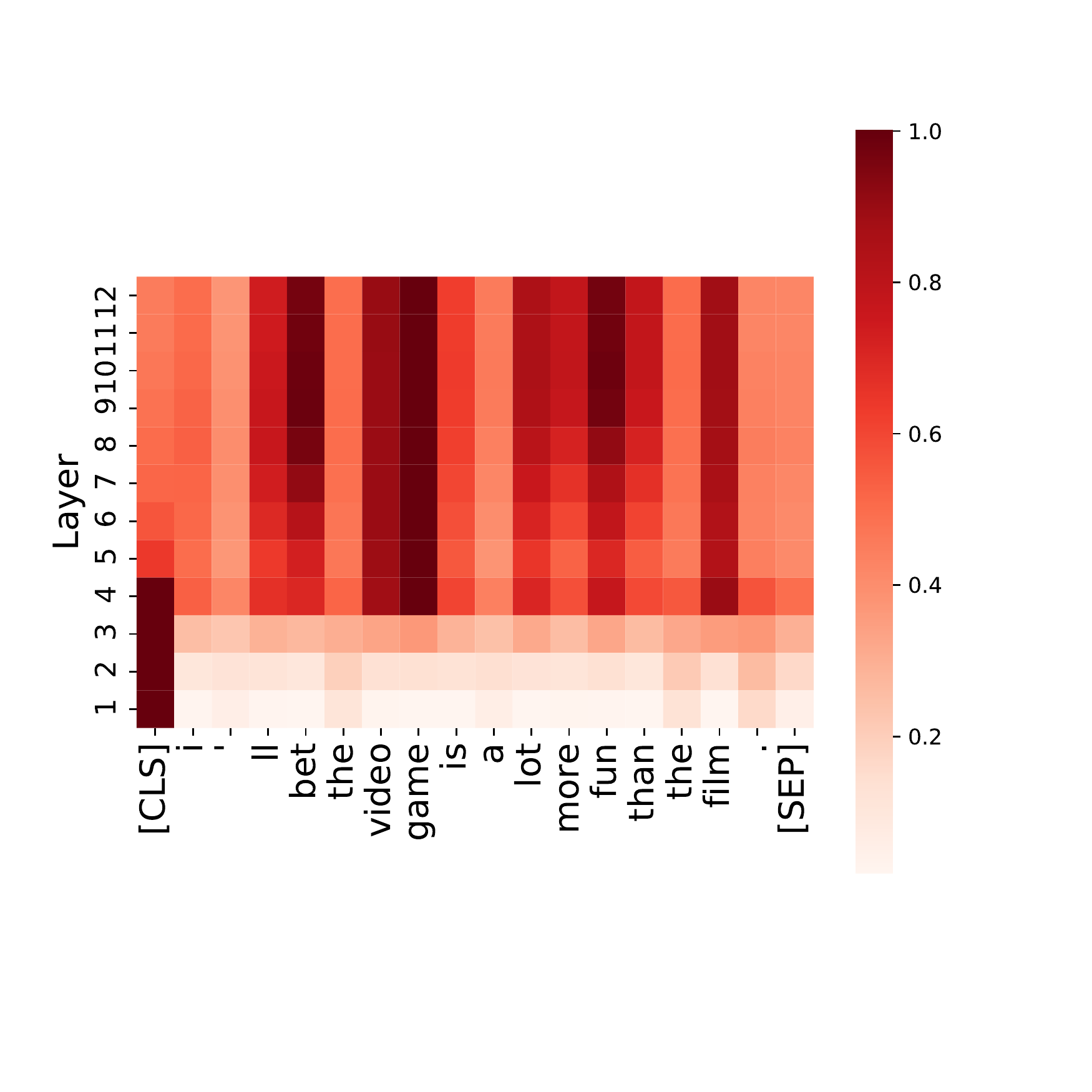}
    }
    \subfloat{
        \includegraphics[width=0.24\textwidth, trim=25 15 50 50, clip] {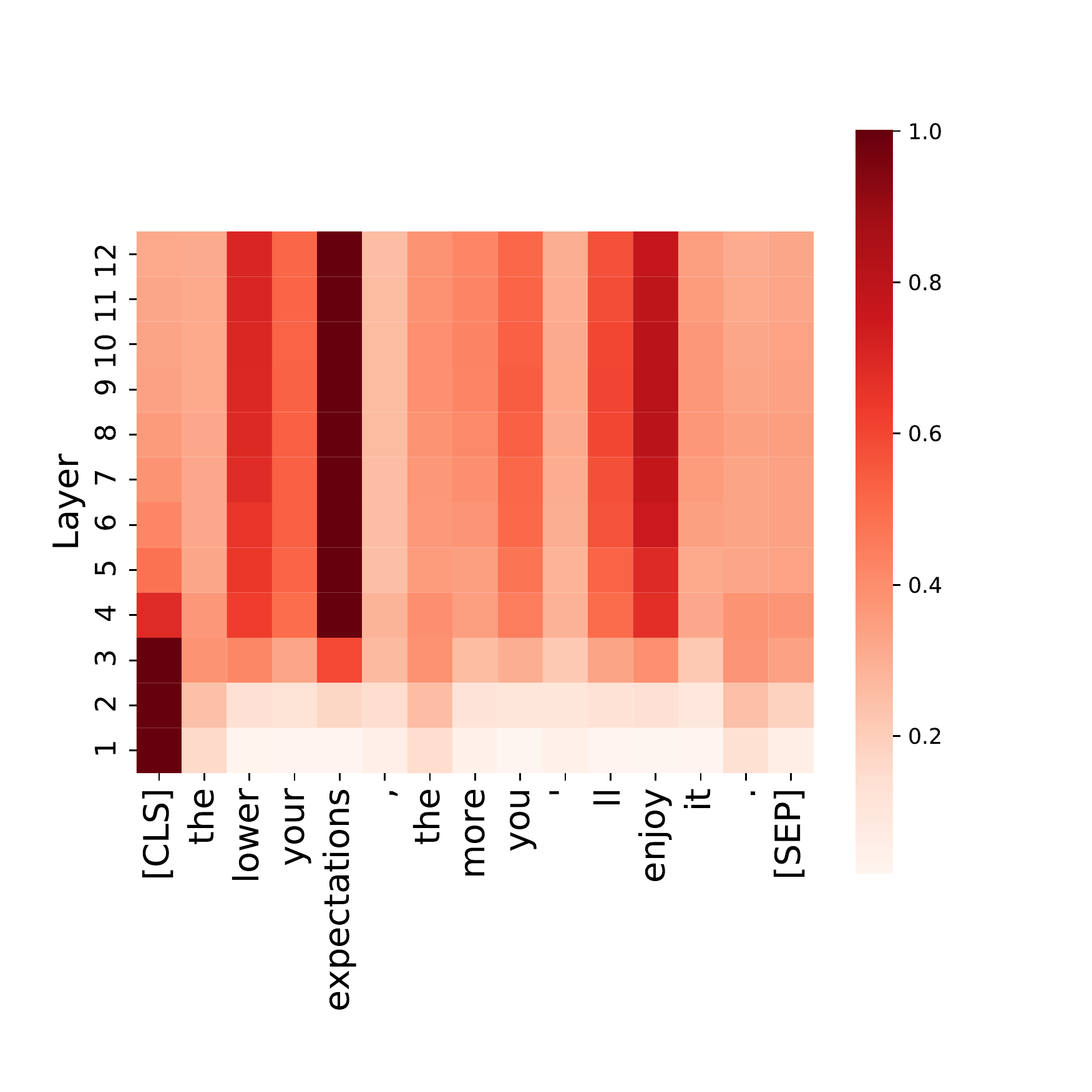}
    }
    \subfloat{
        \includegraphics[width=0.23\textwidth, trim=25 15 50 50, clip] {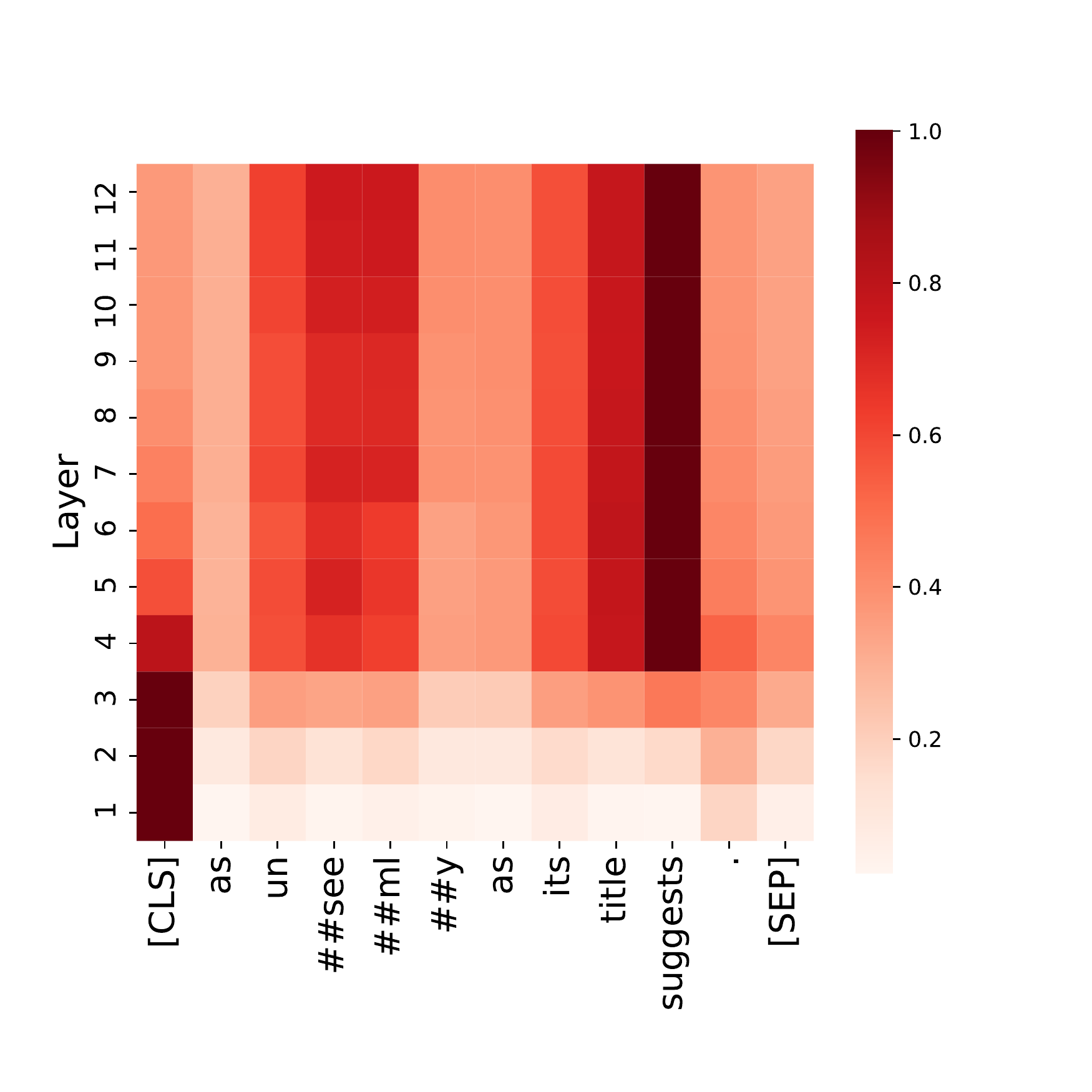}
    }

    \caption{
    Aggregated $\mathcal{N_\textsc{Enc}}$ attribution maps (\methodName) for the \textsc{[CLS]} token for fine-tuned BERT on SST2 dataset (sentiment analysis). These examples were misclassified by the model.
    }
    \label{fig:sst2_errors}
\end{figure*}

\subsection{More Examples}
\label{more_examples}
\begin{figure*}[t]
\centering
    % left bottom right top
    \includegraphics[width=0.98\textwidth, trim=90 220 90 120, clip]{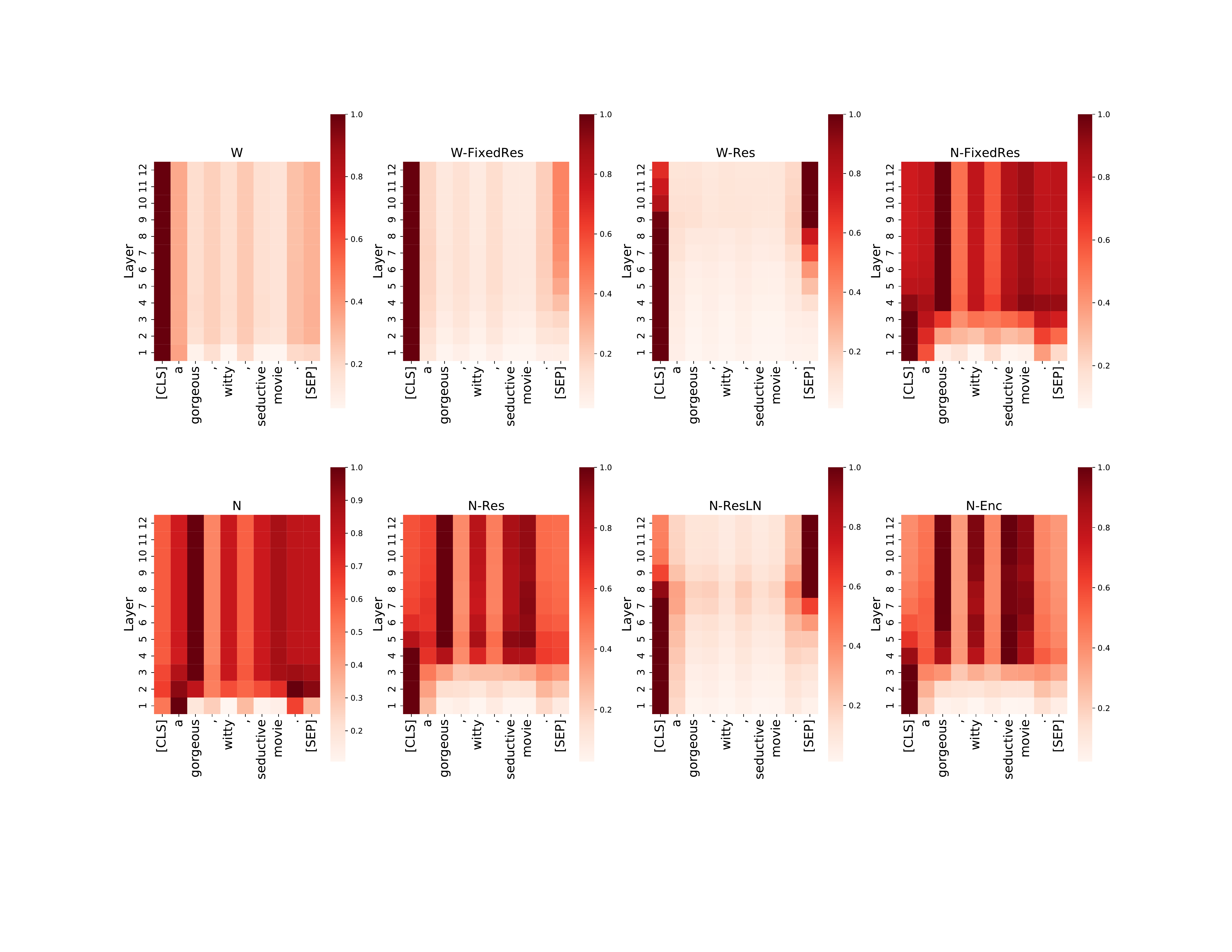}
    \caption{Aggregated attributions via rollout with different methods across layers. The model is fine-tuned on SST2 dataset and the attention of the CLS token is shown in each layer.}
    \label{fig:sst2_cls_all}
\end{figure*}

Aggregated attributions by different methods throughout layers is shown in Figure~\ref{fig:sst2_cls_all}. Our proposed method shows more plausible results.

\begin{figure*}[t]
\centering
    % left bottom right top
    \includegraphics[width=0.98\textwidth, trim=90 160 90 70, clip]{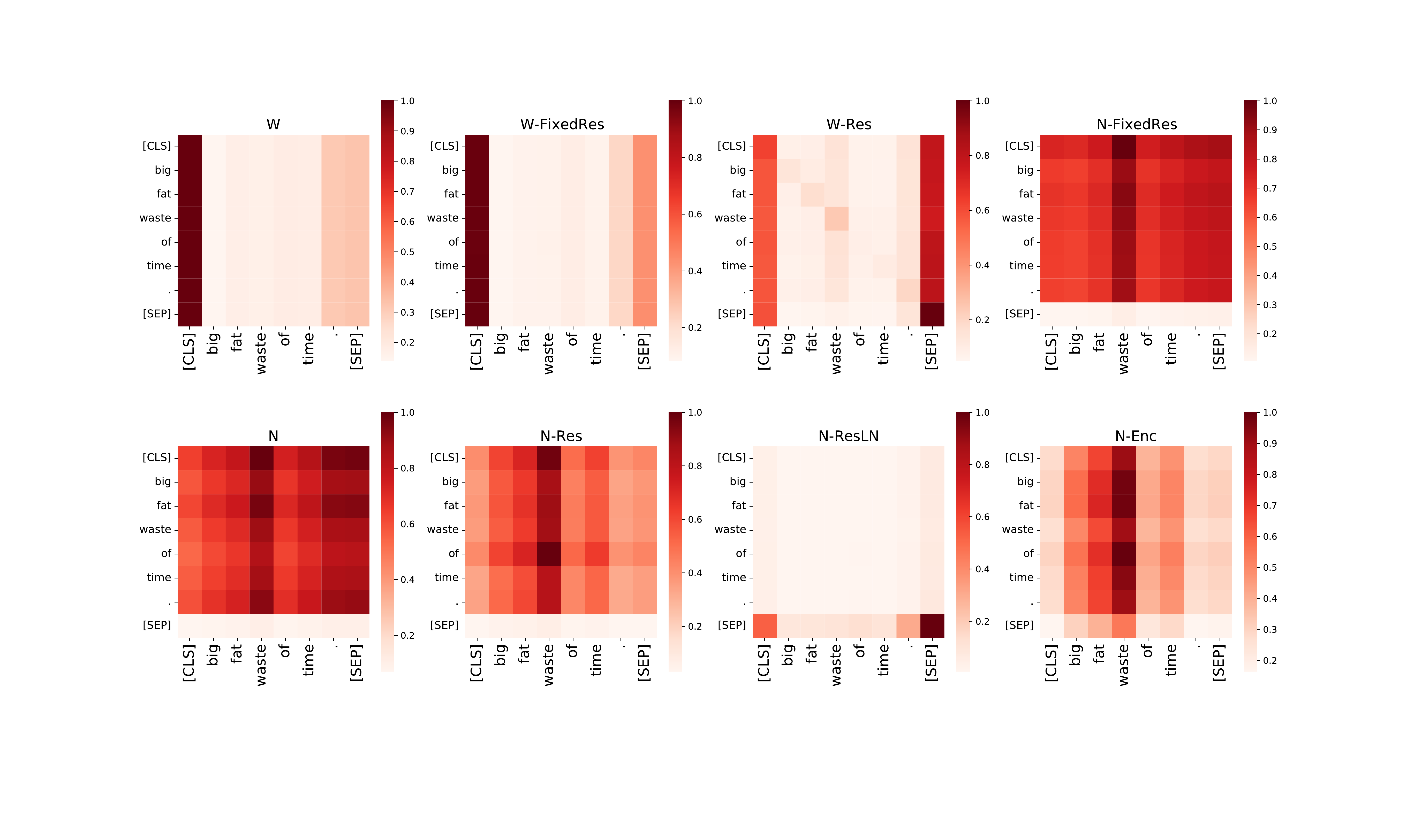}
    \caption{Aggregated attributions via rollout with different methods in layer 12. The model is fine-tuned on SST2 dataset. Each row indicates how much other tokens impact the token written on the row.}
    \label{fig:sst2_tokens}
\end{figure*}

Aggregated attribution map for layer 12 is shown in Figure~\ref{fig:sst2_tokens}. In this figure, the effect of each token can be seen on all other tokens and not just the [CLS] token.

\begin{figure*}[t]
\centering
    % left bottom right top
    \subfloat{
        % left bottom right top
        \includegraphics[width=0.24\textwidth, trim=20 12 50 50, clip] {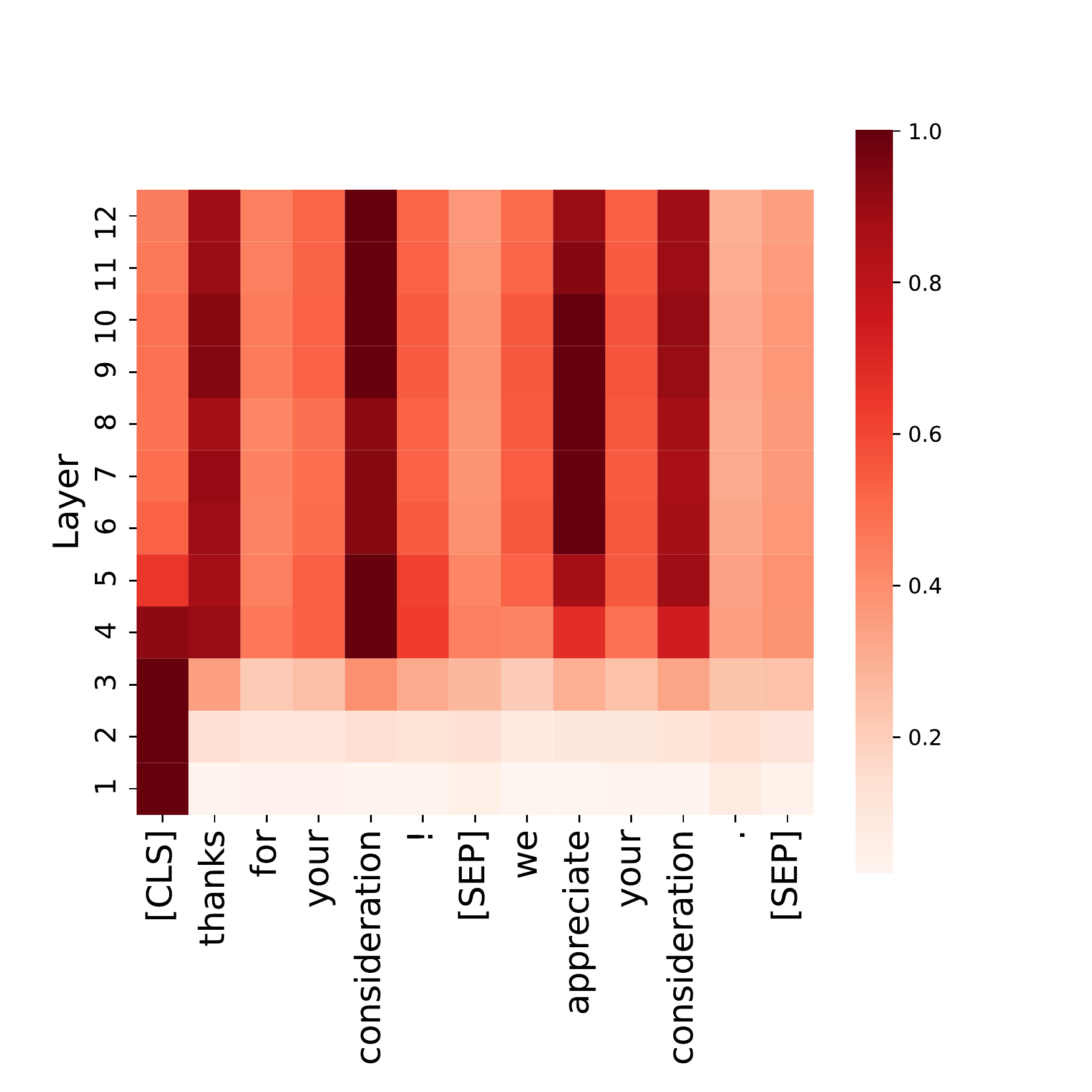}
    }
    \subfloat{
        \includegraphics[width=0.23\textwidth, trim=25 20 50 50, clip] {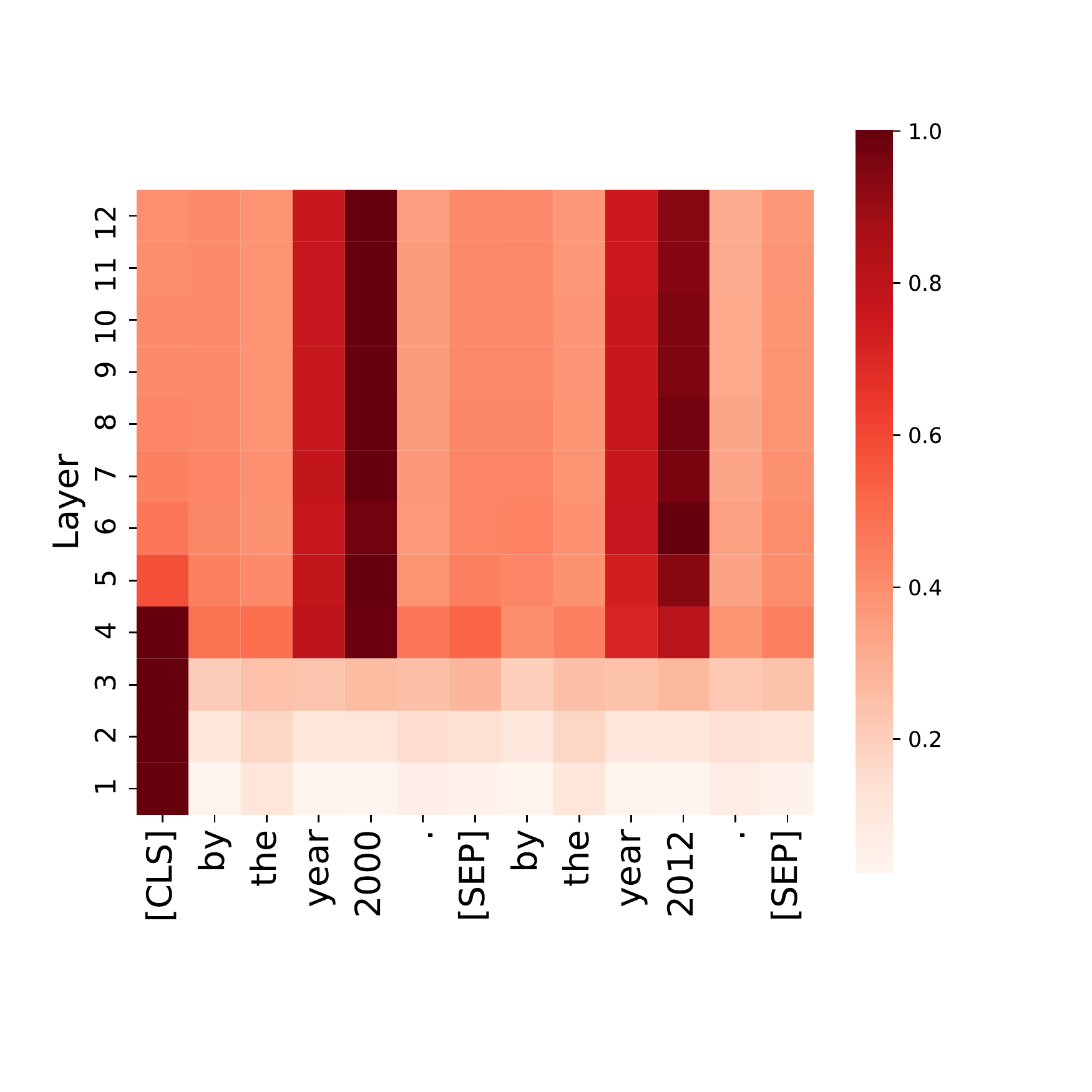}
    }
    \subfloat{
        \includegraphics[width=0.23\textwidth, trim=25 15 50 50, clip] {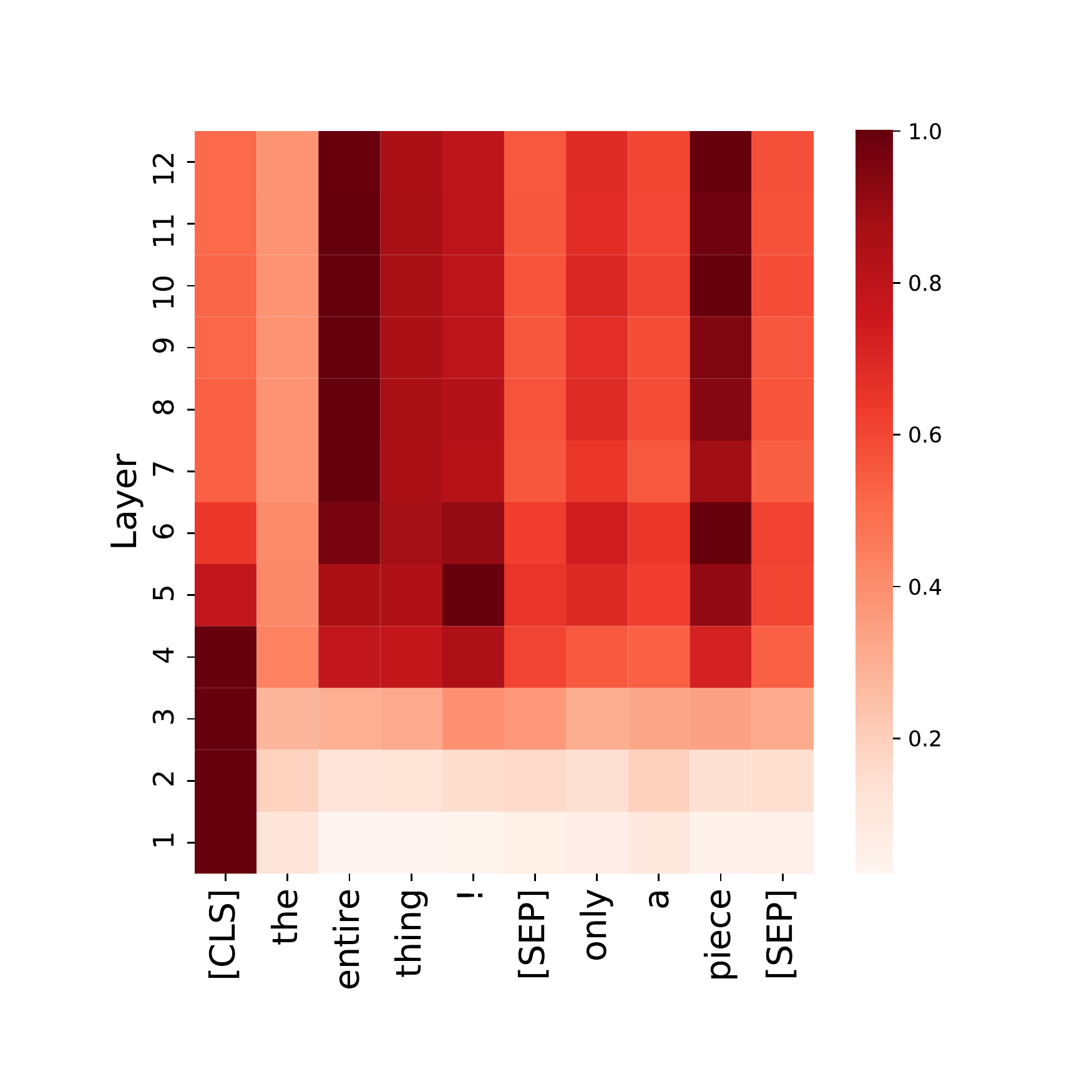}
    }
    \subfloat{
        \includegraphics[width=0.24\textwidth, trim=25 20 50 50, clip] {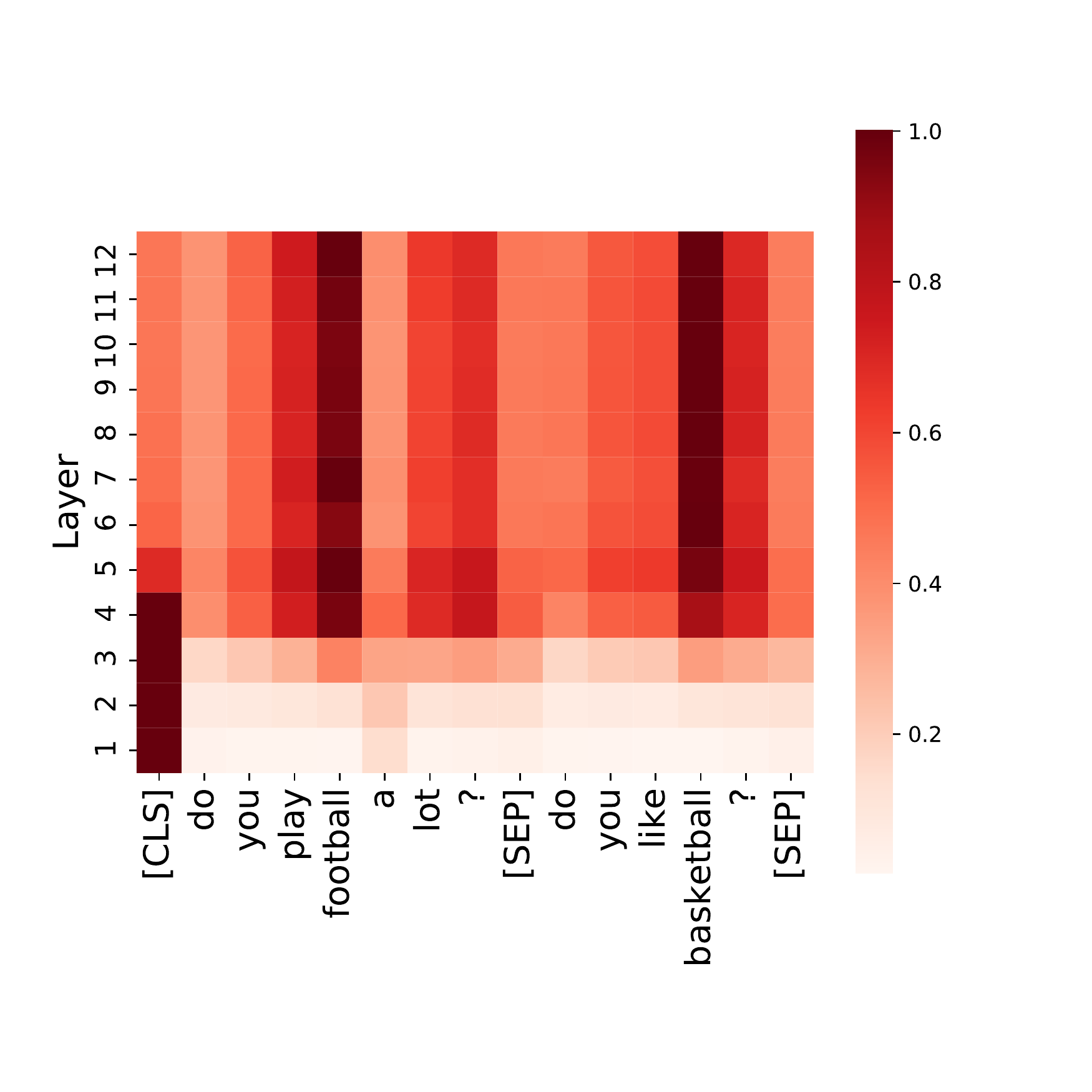}
    }
    \caption{
    Aggregated $\mathcal{N_\textsc{Enc}}$ attribution maps (\methodName) for the \textsc{[CLS]} token for fine-tuned BERT on MNLI dataset.
    }
    \label{fig:mnli_qualitative}
\end{figure*}

\begin{figure*}[t]
\centering
    % left bottom right top
    \subfloat{
        % left bottom right top
        \includegraphics[width=0.24\textwidth, trim=20 12 50 50, clip] {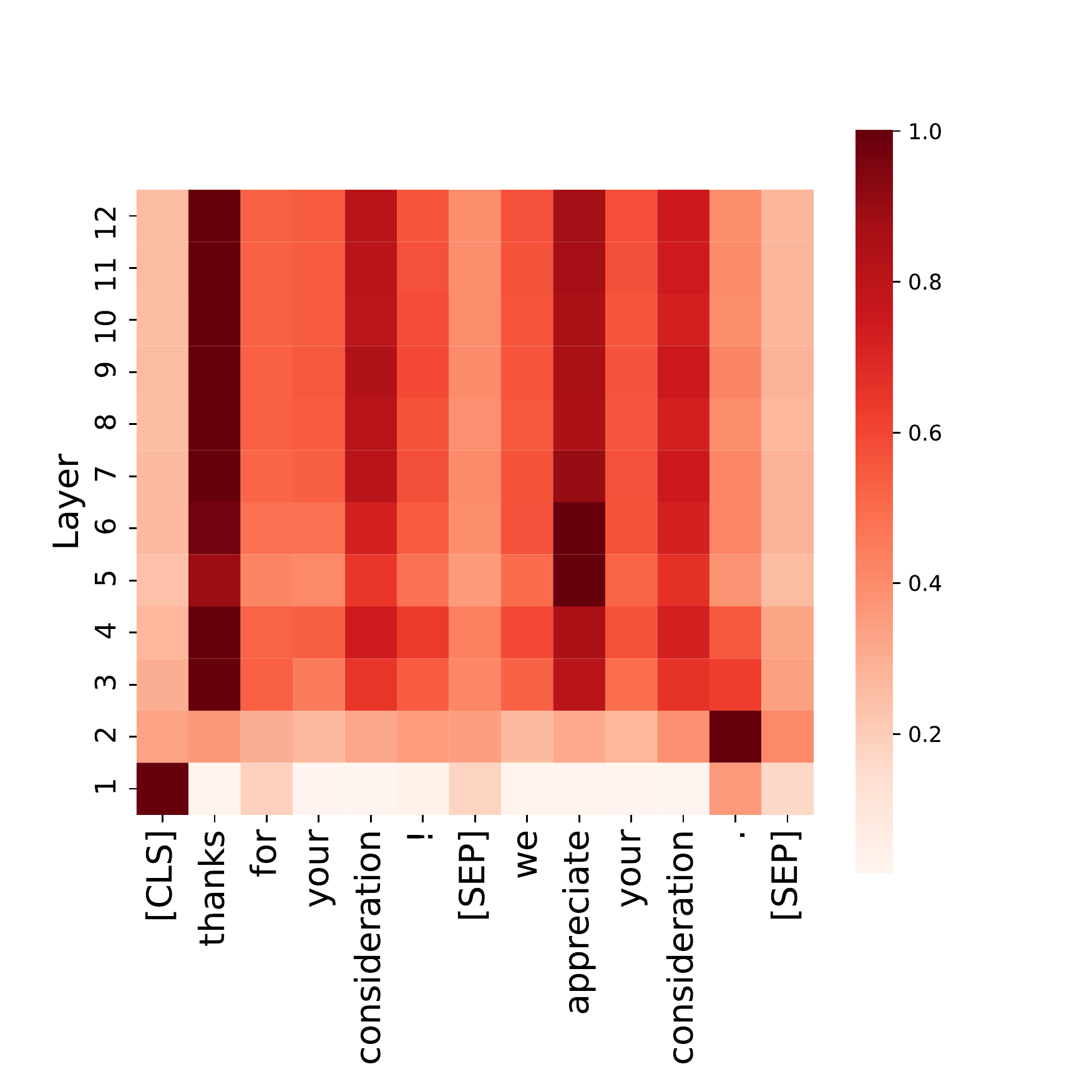}
    }
    \subfloat{
        \includegraphics[width=0.23\textwidth, trim=25 20 50 50, clip] {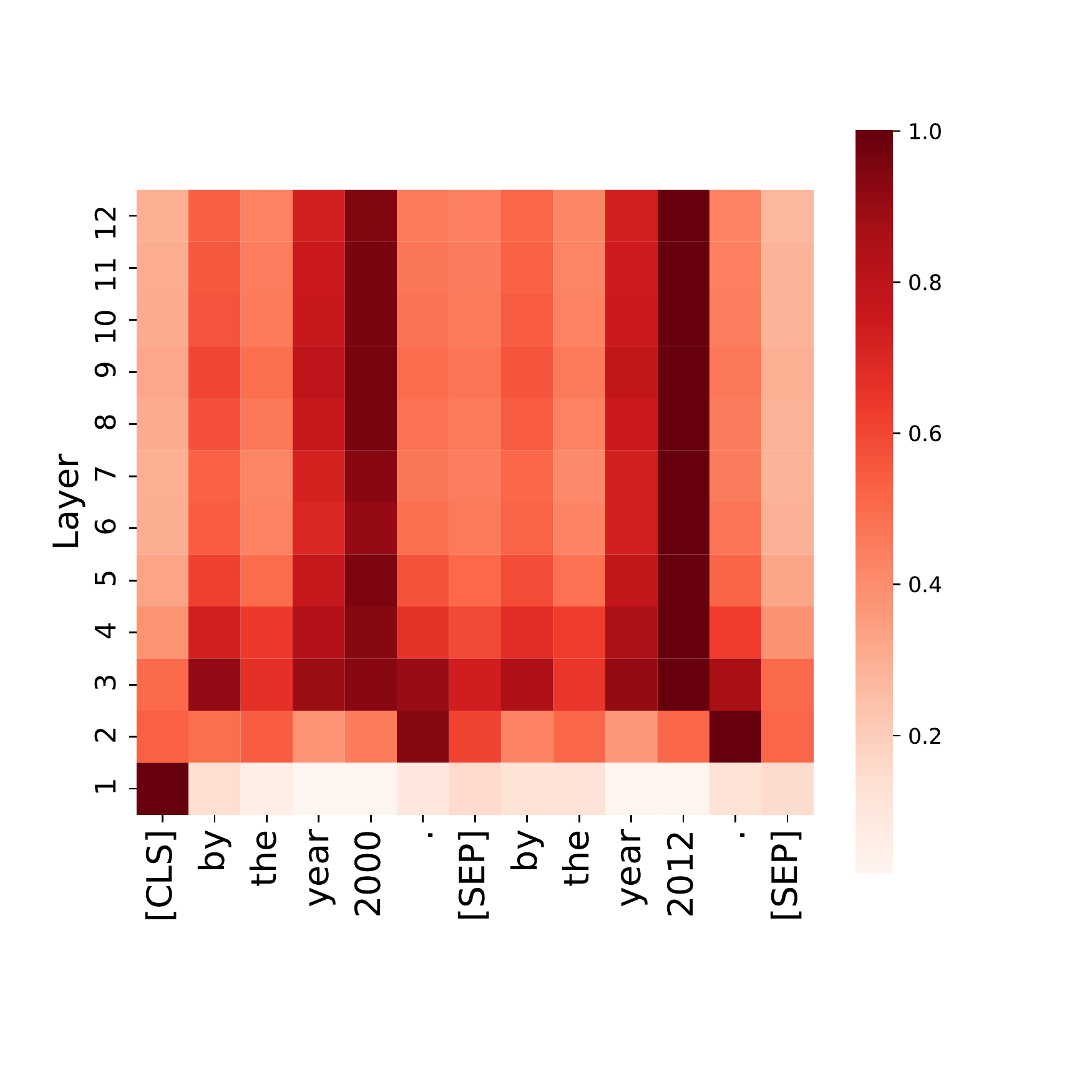}
    }
    \subfloat{
        \includegraphics[width=0.23\textwidth, trim=25 15 50 50, clip] {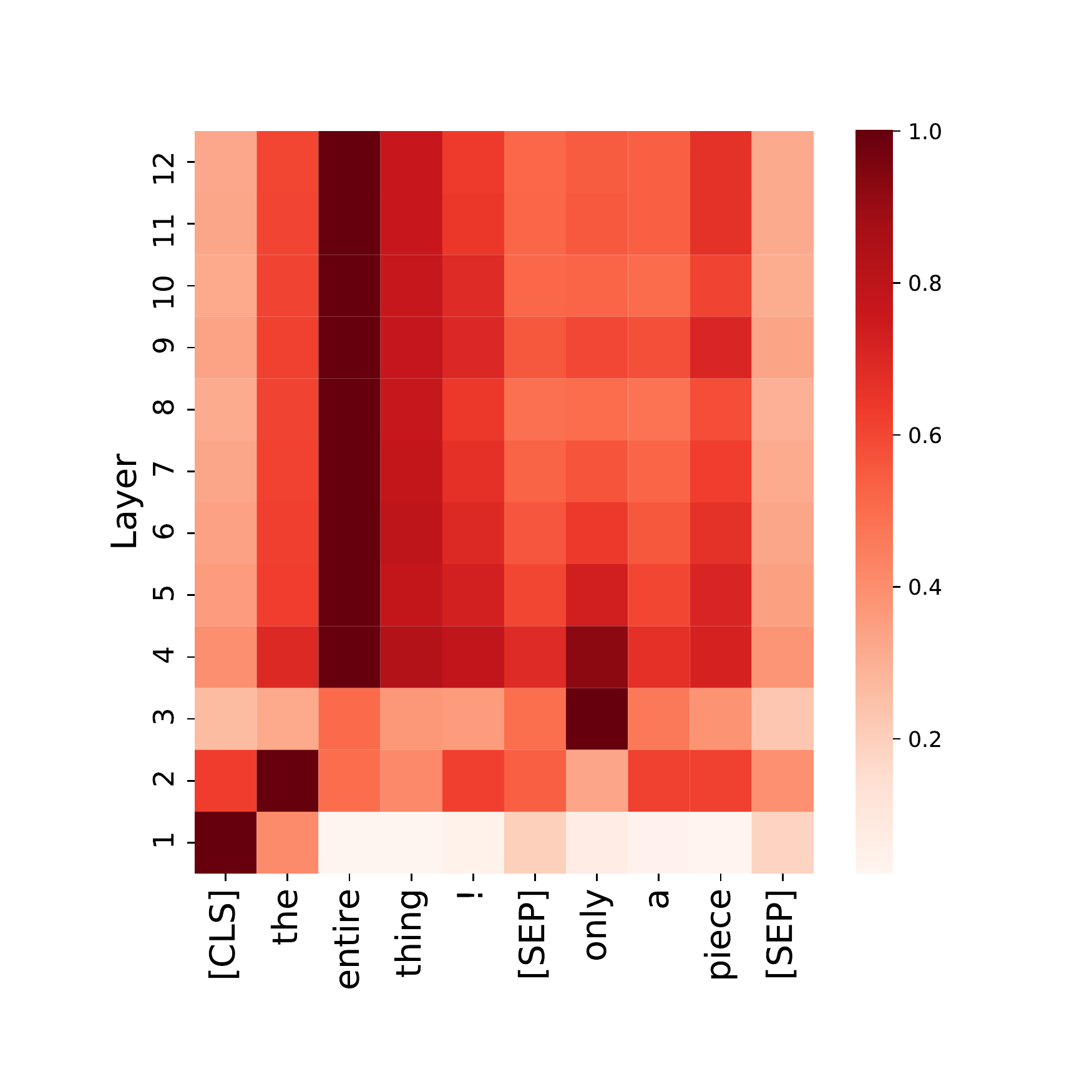}
    }
    \subfloat{
        \includegraphics[width=0.24\textwidth, trim=25 20 50 50, clip] {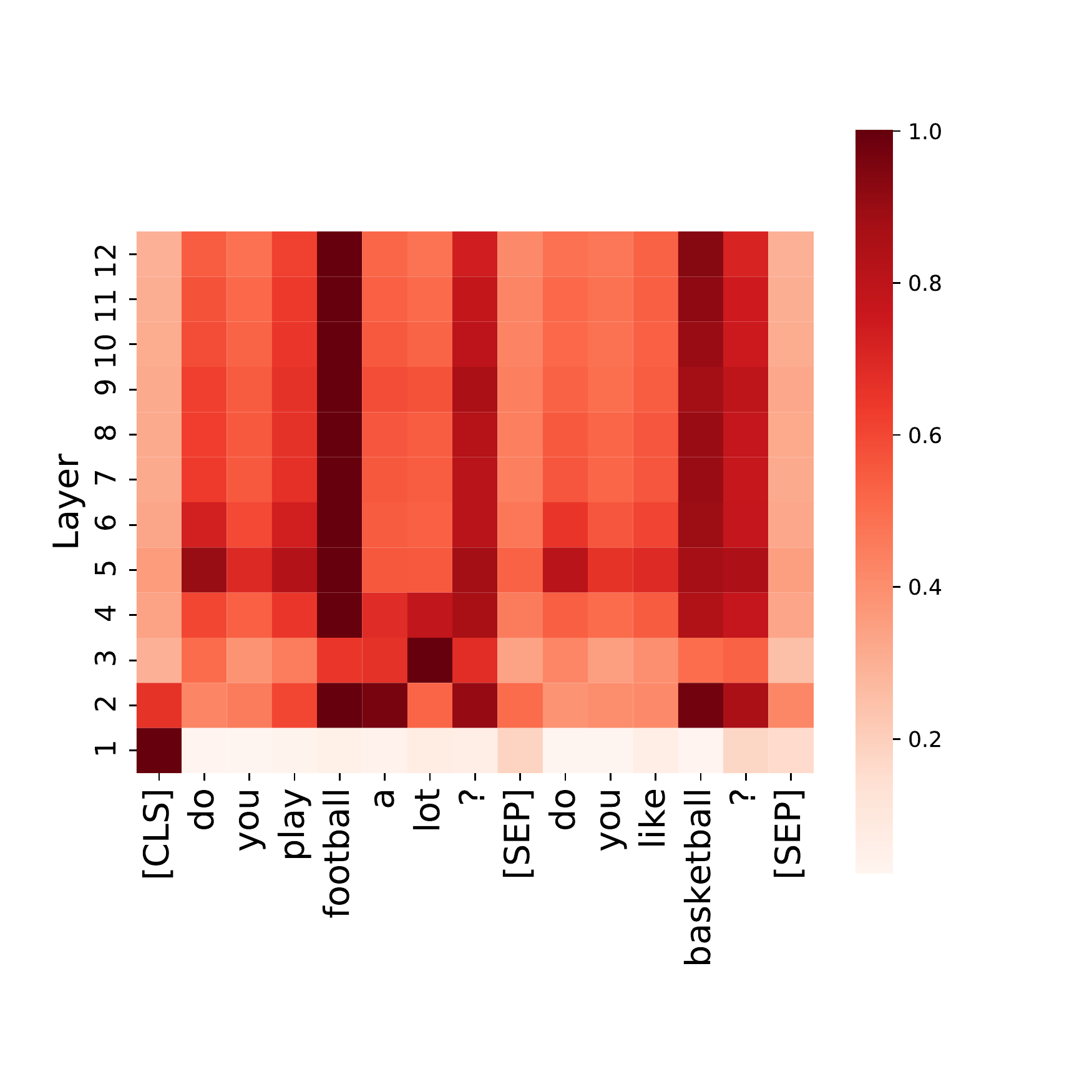}
    }
    \caption{
    Aggregated $\mathcal{N_\textsc{Enc}}$ attribution maps (\methodName) for the \textsc{[CLS]} token for fine-tuned ELECTRA on MNLI dataset.
    }
    \label{fig:mnli_qualitative_electra}
\end{figure*}

\begin{figure*}[t]
\centering
    % left bottom right top
    \subfloat{
        % left bottom right top
        \includegraphics[width=0.24\textwidth, trim=100 12 50 50, clip] {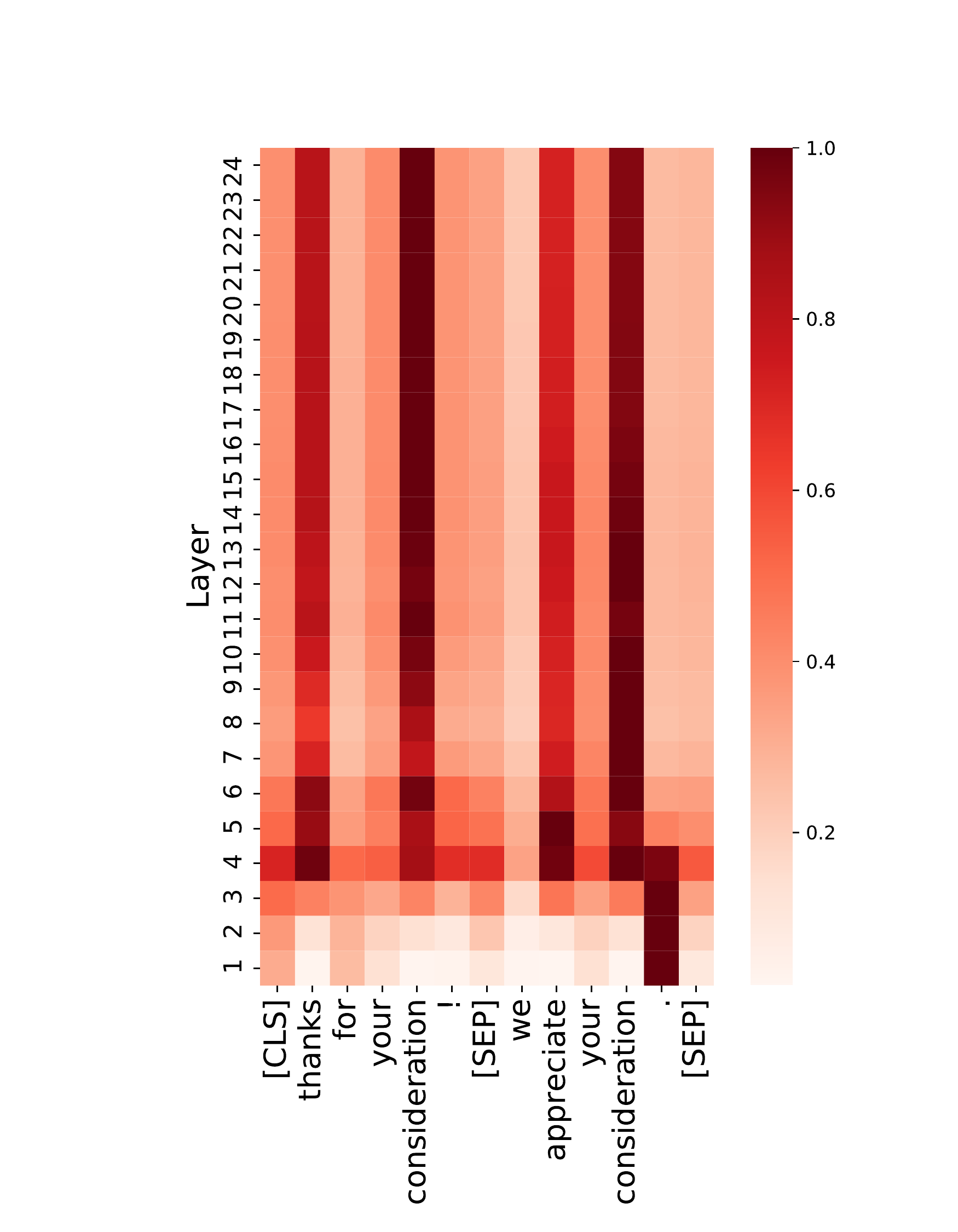}
    }
    \subfloat{
        \includegraphics[width=0.23\textwidth, trim=100 20 50 50, clip] {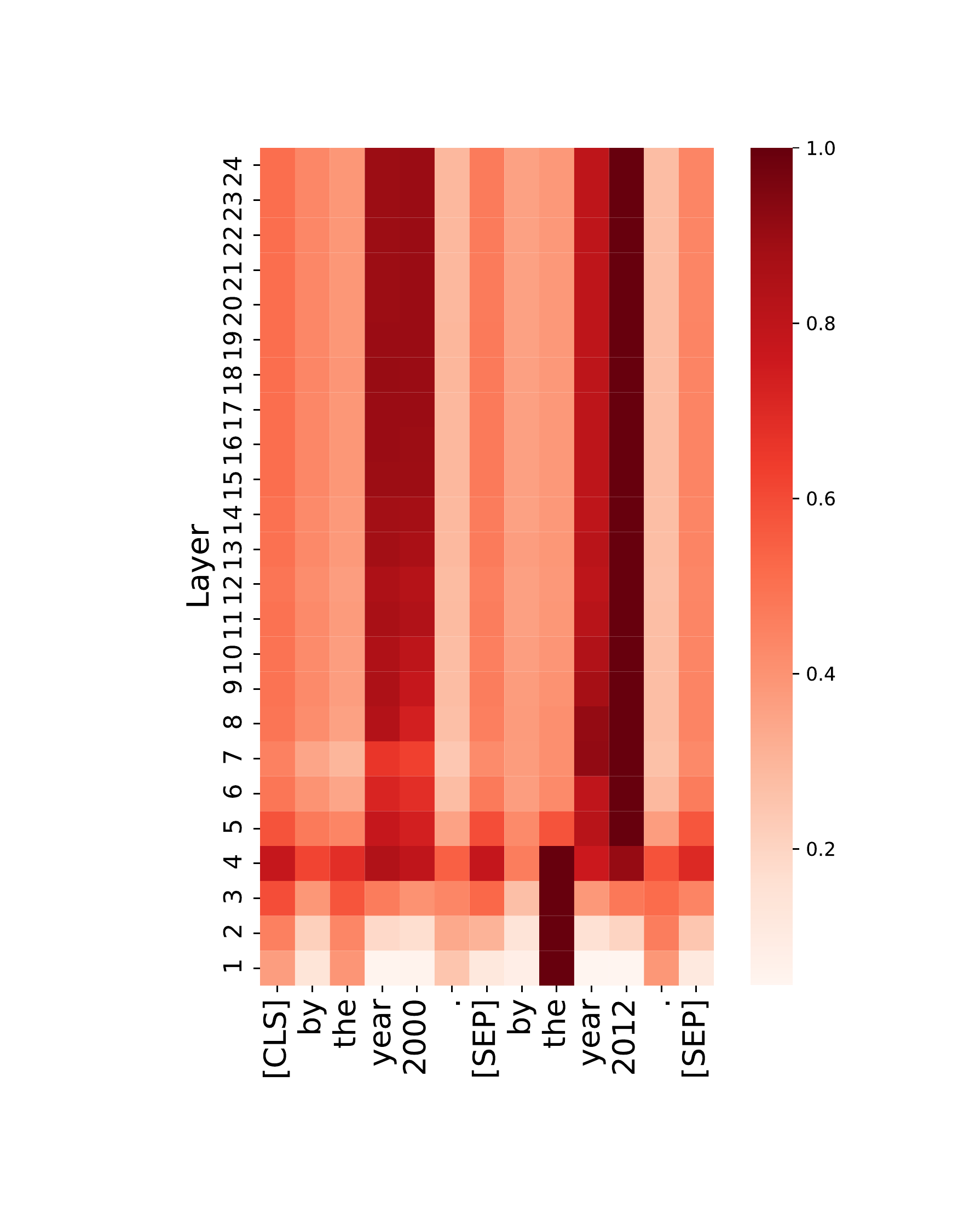}
    }
    \subfloat{
        \includegraphics[width=0.23\textwidth, trim=100 15 50 50, clip] {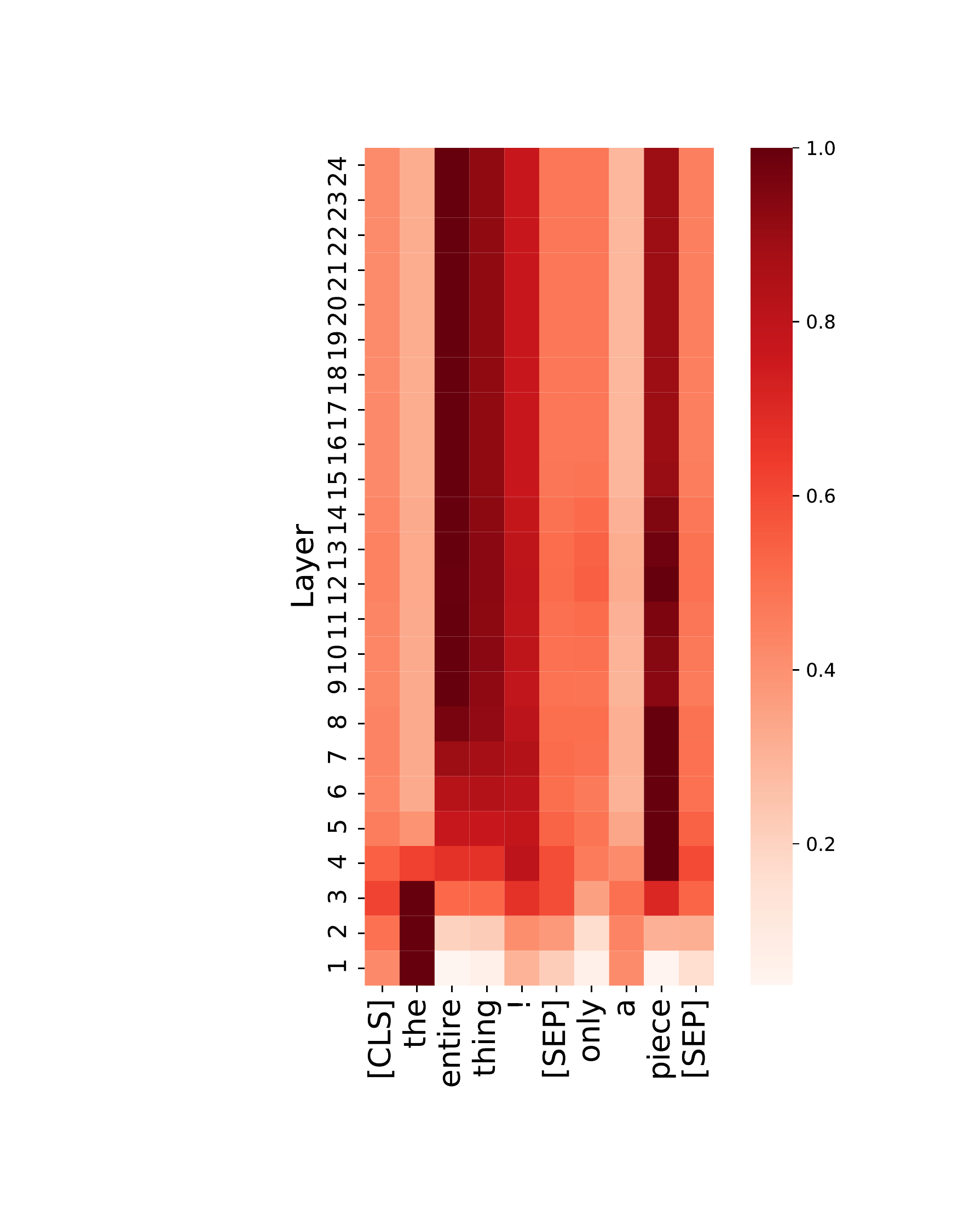}
    }
    \subfloat{
        \includegraphics[width=0.24\textwidth, trim=50 20 50 50, clip] {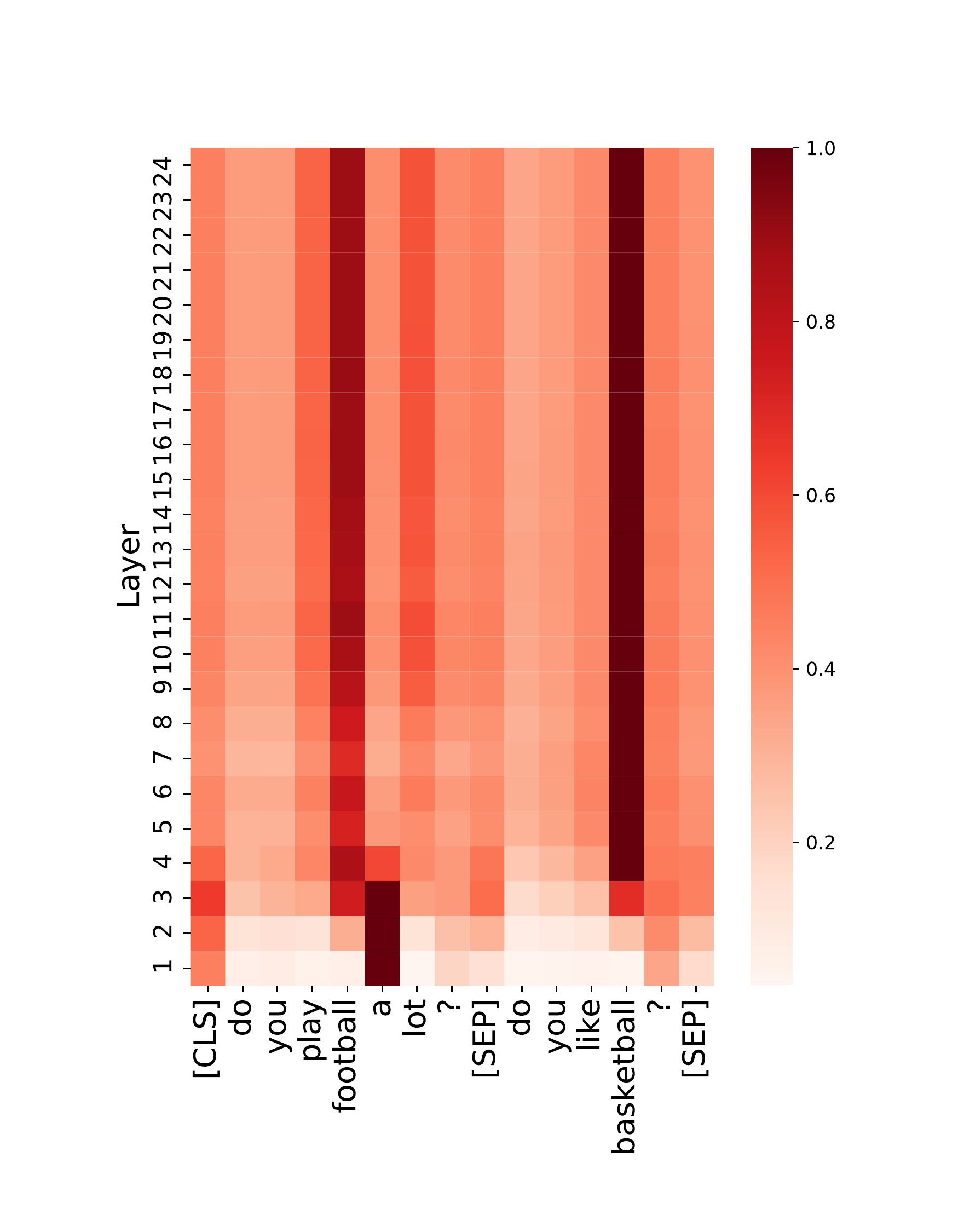}
    }
    \caption{
    Aggregated $\mathcal{N_\textsc{Enc}}$ attribution maps (\methodName) for the \textsc{[CLS]} token for fine-tuned BERT-large on MNLI dataset.
    }
    \label{fig:mnli_qualitative_bert_large}
\end{figure*}

More examples for MNLI dataset are shown for BERT-base in Figure~\ref{fig:mnli_qualitative}, for BERT-large in Figure~\ref{fig:mnli_qualitative_bert_large}, and for ELECTRA in Figure~\ref{fig:mnli_qualitative_electra}.
Moreover, misclassified examples of SST2 dataset are shown in Figure~\ref{fig:sst2_errors}.


\end{document}